\documentclass{article}

\usepackage[final]{neurips_2022}

\setcitestyle{numbers,square,comma} %Citation-related commands
\usepackage[utf8]{inputenc} % allow utf-8 input
\usepackage[T1]{fontenc}    % use 8-bit T1 fonts
\usepackage{xcolor}
\usepackage[pagebackref=true,breaklinks=true,colorlinks,citecolor=blue,urlcolor=blue,linkcolor=black,bookmarks=false]{hyperref}
\usepackage{booktabs}       % professional-quality tables
\usepackage{amsfonts}       % blackboard math symbols
\usepackage{nicefrac}       % compact symbols for 1/2, etc.
\usepackage{microtype}      % microtypography
\usepackage{enumitem}
\usepackage{pifont}
\usepackage{url}
\usepackage{graphicx}
\usepackage{amsmath}
\usepackage{amsfonts}
\usepackage{amssymb}
\usepackage{bbm}
\usepackage{booktabs}
\usepackage{siunitx}
\usepackage{changepage}
\usepackage{multirow}
\usepackage{wrapfig, lipsum}
\usepackage{makecell}
\usepackage{xspace}
\usepackage{dashrule}
\DeclareMathOperator*{\argmax}{arg\,max}
\DeclareMathOperator*{\argmin}{arg\,min}

\newcommand{\method}[0]{\textsc{VisFIS}\xspace}

\newcommand{\replace}[0]{$\mathrm{Replace}$\xspace}
\newcommand{\replaced}[0]{$\mathrm{Replaced}$\xspace}

\title{\textsc{VisFIS}: Visual Feature Importance Supervision \\ with Right-for-the-Right-Reason Objectives}

\author{
Zhuofan Ying,\thanks{Equal contribution.} \; Peter Hase,\footnotemark[1] \; {\normalfont and} Mohit Bansal  \\
  Department of Computer Science\\
  University of North Carolina at Chapel Hill \\
  \texttt{\{zfying, peter, mbansal\}@cs.unc.edu} \\
}

\begin{document}

\maketitle

\begin{abstract}
Many past works aim to improve visual reasoning in models by supervising feature importance (estimated by model explanation techniques) with human annotations such as highlights of important image regions. 
However, recent work has shown that performance gains from feature importance (FI) supervision for Visual Question Answering (VQA) tasks persist even with \emph{random} supervision, suggesting that these methods do not meaningfully align model FI with human FI. 
In this paper, we show that model FI supervision can meaningfully improve VQA model accuracy as well as performance on several Right-for-the-Right-Reason (RRR) metrics by optimizing for four key model objectives: (1) accurate predictions given limited but sufficient information (Sufficiency); (2) max-entropy predictions given no important information (Uncertainty); (3) invariance of predictions to changes in unimportant features (Invariance); and (4) alignment between model FI explanations and human FI explanations (Plausibility). 
Our best performing method, Visual Feature Importance Supervision (\method), outperforms strong baselines on benchmark VQA datasets in terms of both in-distribution and out-of-distribution accuracy.
While past work suggests that the mechanism for improved accuracy is through improved explanation plausibility,
we show that this relationship depends crucially on explanation \emph{faithfulness} (whether explanations truly represent the model's internal reasoning). 
Predictions are more accurate when explanations are plausible \emph{and} faithful, and not when they are plausible but not faithful. 
Lastly, we show that, surprisingly, RRR metrics are not predictive of out-of-distribution model accuracy when controlling for a model's in-distribution accuracy, which calls into question the value of these metrics for evaluating model reasoning.\footnote{All supporting code for experiments in this paper is available at \url{https://github.com/zfying/visfis}.}
  
\end{abstract}

\section{Introduction}
\label{sec:introduction}

Many past works aim to teach models to ignore spurious features by making use of additional information about which features in an input are important \cite{kulesza2015principles, ross2017right, teso2019explanatory}. 
For example, individual words can be annotated as (un)important in NLP tasks \cite{ross2017right, zhong2019fine}, or regions of image pixels can be highlighted by humans as extra supervision for vision tasks \cite{das2017human, singla2022core}.
In this broad class of feature importance (FI) supervision methods, human annotations of important features are typically provided for individual datapoints, and methods often use data augmentation or gradient supervision to encourage models to rely only on important features when making predictions.
Such approaches have seen performance improvements in image classification \cite{simpson2019gradmask, chang2021towards}, text classification \cite{liu2019incorporating, zhong2019fine, teney2020learning}, and multimodal visual question answering (VQA) tasks \cite{selvaraju2019taking, wu2019self, liu2022answer}. 

\begin{figure}[t]
  \begin{center}
    \includegraphics[width=.98\textwidth]{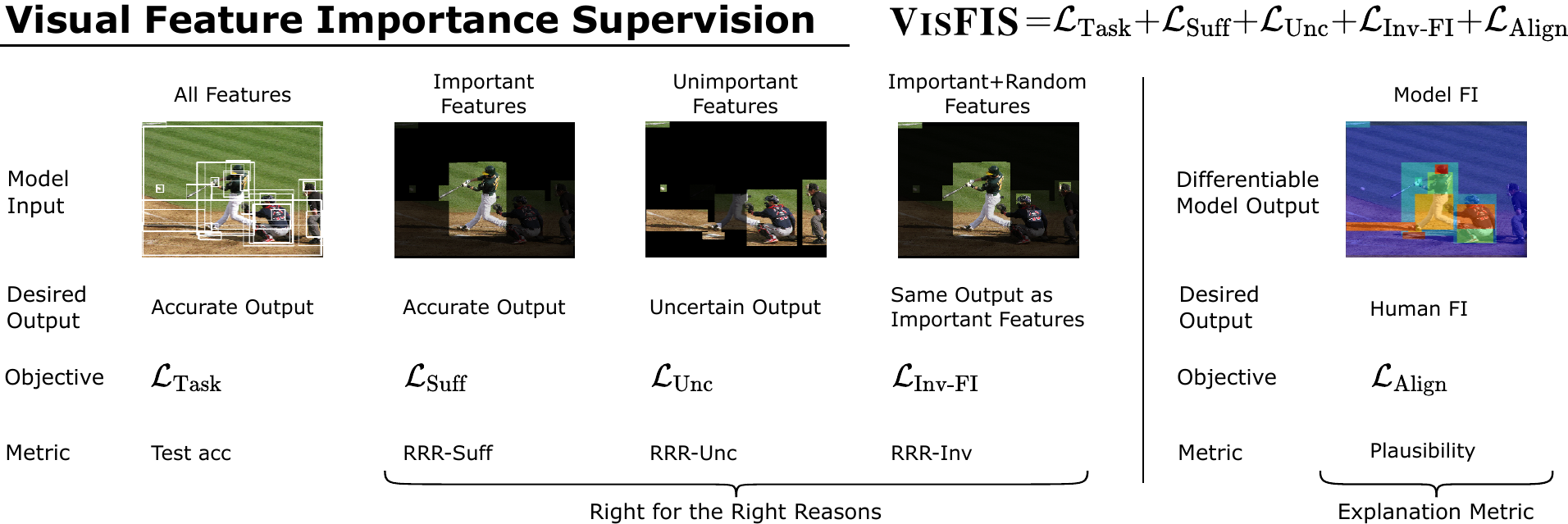}
  \end{center}
  \vspace{-5pt}
  \caption{We depict five core desiderata for VQA models with associated metrics and objectives. We seek models that (1) are accurate given the full visual input, (2) are accurate given only important features, (3) are appropriately uncertain given only unimportant features, (4) are invariant to unimportant features, and (5) yield FI estimates (explanations) that align with human FI. \method combines the five objectives.
  Note objectives \#2--\#5 require additional FI annotations.
  Darkened image regions correspond to bounding box representations that have been $\mathrm{Replaced}$ (see Sec. \ref{sec:background_and_notation}). 
  }
  \label{fig:main-fig}
  \vspace{-11pt}
\end{figure}

One of the primary motivations behind these approaches is to improve task accuracy by making models ``Right for the Right Reasons'' \cite{ross2017right}.
While existing FI supervision methods for VQA models can improve accuracy \cite{selvaraju2019taking, wu2019self}, recent work has cast doubt on whether human supervision is the source of these improvements. 
Specifically, \citet{shrestha-etal-2020-negative} show that VQA improvements persist even when \emph{random image FI annotations} are used as supervision, suggesting that existing approaches may not extract meaningful signal from the human annotations.
Motivated by this shortcoming, we explore several aspects of the FI supervision problem in the context of VQA tasks:

\vspace{2pt} 
\textbf{Improving VQA accuracy with FI supervision via four key model objectives (Sec. \ref{sec:supervision_improve_accuracy}).} 
Past VQA methods focus on data augmentation techniques \cite{liu2022answer} or ways to directly supervise model feature importance \cite{selvaraju2019taking, wu2019self}. 
We make use of four key objectives (represented in Fig. \ref{fig:main-fig}): (1) a Sufficiency objective encouraging the model to predict the correct label given only important input features \cite{chang2021towards}; (2) an Uncertainty objective encouraging max-entropy outputs when given only unimportant features; (3) an Invariance objective encouraging model outputs to be invariant to changes in unimportant features \cite{ross2017right}; and (4) an Align objective that penalizes the model when its FI estimates differ from human FI annotations \cite{selvaraju2019taking}.
Our best performing method, termed Visual Feature Importance Supervision (\method), combines these strategies strategies in a novel manner, improving both in-distribution and out-of-distribution accuracy. 
Following guidelines from \citet{shrestha-etal-2020-negative}, we show that this improvement does not occur with random supervision, meaning \method learns from human supervision itself. 
Lastly, after analyzing how explanation plausibility and faithfulness \cite{jacovi2020towards} relate to accuracy at the datapoint level, we suggest that FI supervision improves prediction accuracy by improving the plausibility of \emph{faithful} FI explanations, rather than plausibility alone as past work has suggested \cite{selvaraju2019taking, chang2021towards}.

\vspace{2pt} 
\textbf{Evaluating models on new Right for the Right Reason (RRR) metrics (Sec. \ref{sec:supervision_make_models_RRR}).} Beyond measuring model accuracy, past works evaluate models on a few Right for the Right Reason metrics in order to understand whether model reasoning is correct \cite{selvaraju2019taking, shrestha-etal-2020-negative, yang2020object, chang2021towards, Gupta_2022_CVPR, friedrich2022typology}. 
Correct reasoning is valuable because it suggests that models will generalize to test data that we might not be able to verify their performance on, which can occur e.g. when there are exponentially many cases we wish to test or when such data is prohibitively expensive to collect. 
We propose a broad set of RRR metrics for model evaluation, with similar motivation to our key model objectives above (see Fig. \ref{fig:main-fig}). 
Specifically, in addition to measuring existing metrics for (1) in-distribution accuracy, (2) out-of-distribution accuracy \cite{agrawal2018vqacp,clark2019don-LMH,cadene2019rubi, chen2020counterfactual, chang2021towards}, and (3) model accuracy on sufficient feature subsets (RRR-Suff) \cite{chang2021towards}, 
we also evaluate (4) model uncertainty given uninformative inputs (RRR-Unc), and (5) model invariance to changes in unimportant features (RRR-Inv). 
These metrics help verify that models can: arrive at correct answers relying only on features that are actually important (metric \#3), are invariant to the addition or removal of unimportant features that should not affect the label (metric \#4), and are appropriately uncertain about the model class when the input contains no meaningful evidence for any class (metric \#5).

\vspace{2pt} 
\textbf{Predicting model generalization to OOD data with RRR metrics (Sec. \ref{sec:predict_OOD_generalization}).} The practical value of the above RRR metrics can be considered in terms of their ability to inform us about model performance on data that we are not able to test on. We simulate this situation by evaluating models on in-distribution (ID) data and predicting whether the models will generalize to OOD data based on their accuracy and RRR metrics for ID data. 
Surprisingly, we find that both existing RRR metrics and our new ones do not better predict OOD accuracy than ID accuracy does on its own. This finding suggests that these metrics may not be a good evaluation of model reasoning, and that there is no good replacement yet for evaluating model accuracy on OOD data in addition to ID data.

In summary of our contributions, we show that:
\begin{enumerate}[itemsep=0pt, wide=0pt, leftmargin=*, after=\strut]
\item FI supervision can improve both ID and OOD model accuracy on several benchmark VQA datasets. In particular, \method improves over unsupervised baselines and the previous state-of-the-art on CLEVR-XAI by up to 4.7 points on OOD data.

\item Explanation plausibility correlates with model accuracy only when explanations are also faithful, which sheds light on the mechanism by which FI supervision improves model accuracy.

\item FI supervision improves model performance on several RRR metrics, including new invariance and uncertainty metrics. 

\item RRR metrics do \emph{not} correlate better with OOD accuracy than ID model accuracy does on its own. Consequently, RRR metrics may not be as valuable as previously thought.
\end{enumerate}

\section{Related Work}
\label{sec:related_work}

\looseness=-1
\textbf{Supervising FI explanations.} Past works primarily supervise gradient-based \cite{ross2017right} or attention-based model explanations \cite{zhong2019fine, stacey2022supervising, gao2022aligning, chrysostomou2021enjoy}.
For example, \citet{ross2017right} enforce an $\ell_2$ norm on the gradient of the loss w.r.t. the model input for features marked as unimportant by a human FI explanation. This method appears in several later works \cite{simpson2019gradmask, ghaeini2019saliency, singla2022core}. 
In a VQA setting, \citet{selvaraju2019taking} and \citet{wu2019self} align the entire input gradient with human FI. 
In addition to using input gradients (termed Vanilla Gradient), we consider the Expected Gradients method \cite{erion2021improving}, a computationally efficient implementation of Integrated Gradients \cite{sundararajan_2017_axiomatic}.
Omission or perturbation-based approaches have seen more limited use. \citet{kennedy2020contextualizing} regularize omission-based FI toward 0 for group identifiers in hate speech detection. In addition to simple leave-one-out \cite{li_understanding_2016} and keep-one-in methods, we propose a differentiable version of the popular linear method, SHAP \cite{lundberg_unified_2017}.
For a survey of methods we refer readers to \citet{friedrich2022typology}.
Following the analysis of \citet{shrestha-etal-2020-negative}, we use a random supervision baseline to show that \method succeeds by virtue of additional supervision and not simply via model regularization.

\looseness=-1
\vspace{2pt}
\textbf{Supervised data augmentation.} This line of work uses human explanations to guide data augmentation, sometimes in a human-in-the-loop manner. For instance, \citet{teney2020learning} present LIME explanations to people and solicit feedback that is converted into counterexamples for model training. 
\citet{liang2020alice} use expert natural language counterfactual explanations to manufacture new labeled inputs.

\looseness=-1
We build on prior work for our data augmentation objectives. 
Similar to \citet{chang2021towards} and concurrent work from \citet{singla2022core}, our Sufficiency objective encourages models to be accurate given inputs with sufficient features selected according to human FI. 
Our Uncertainty objective reduces model confidence when no important features are provided, while the most related objectives from \citet{chang2021towards} and \citet{liu2022answer} encourage a \emph{different} answer rather than an uncertain output. 
We are not aware of objectives encouraging invariance to changes in unimportant features as our invariance objective does. 
Other concurrent work encourages models to always predict the true label even when unimportant features are swapped with other unimportant features from the data distribution \cite{Gupta_2022_CVPR}.

\vspace{2pt}
\textbf{Right for the Right Reason metrics.} 
Only a few past works explicitly evaluate RRR metrics in addition to test set accuracy. Our metrics include the existing RRR-Suff metric \cite{chang2021towards} and our RRR-Inv and RRR-Unc metrics. 
While explanation plausibility is regularly proposed as an RRR metric \cite{selvaraju2019taking, shrestha-etal-2020-negative, yang2020object, prasad2020extent, friedrich2022typology}, we show that the relationship between accuracy and plausibility is controlled by explanation faithfulness, meaning that plausibility on its own should not be an RRR metric. Recently, \citet{joshi2022er} propose a number of distinct distribution shifts in text classification for evaluating FI supervision techniques according to model OOD accuracy. We use a ``changing prior'' distribution shift standardly used in past work for VQA \cite{das2017human, teney2020value}, and moreover, the focus of our work is on novel RRR objectives, metrics, and analysis of how supervision improves models.

New machine learning metrics are often justified from first principles or on the basis of a strong correlation with some other good metric, like a human rating \cite{papineni2002bleu}. 
Here, we assess RRR metrics on the basis of their correlation with OOD accuracy, while controlling for model ID accuracy. This means that we measure the correlation between ID and OOD accuracy like in past studies \cite{taori2020measuring, miller2020effect, mania2020classifier, teney2022id}, but we also consider RRR metrics as additional explanatory variables for OOD generalization.

\section{Terminology and Notation}
\label{sec:background_and_notation}

\textbf{FI Explanations.} 
We distinguish between a human explanation $e$ and a model explanation $\tilde{e}$ that is obtained algorithmically to explain how a model arrived at its prediction for some datapoint. 
In this paper, human explanations are real-valued annotations for input features (which are bounding box representations for each of our datasets). The score for each bounding box is an indication of its importance to determining the datapoint label, which could roughly be thought of as an answer to the question, ``why did data point $x$ receive label $y$'' \cite{miller2019explanation}.
For several objectives and metrics, we binarize the explanations, selecting a threshold based on the data distribution (see Appendix \ref{app:data_details}).

$\mathrm{\bf Replace}$ \textbf{ Functions.} Both generating and evaluating model explanations often require ``hiding'' input features from a model. 
In practice, we must replace features with some baseline value \cite{sturmfels2020visualizing}.
One simple and common way to replace features is to use an all-zeros feature \cite{li2016understanding, sundararajan_2017_axiomatic, arras-etal-2019-evaluating}.
We compare among several $\mathrm{Replace}$ functions to find the best function for learning from FI supervision (see Appendix \ref{app:replace_functions}). 
We ultimately select the All-Negative-Ones function, which replaces a bounding box feature vector with the all negative ones vector, $\{-1\}^d$.
We use $x_e$ to denote a version of the input $x$ where features where $e$ is 0 are replaced via our $\mathrm{Replace}$ function. 

\textbf{Model Notation.} We parametrize a distribution $p(y|x)=f_\theta(x)$ for classification purposes. Here, $\hat{y}=\argmax_y f_\theta(x)_y$ is the model prediction, $f_\theta(x)_{\hat{y}}$ is the predicted probability, and $\mathcal{Y}$ is the space of eligible answers, which is a large set that is shared across all questions in our VQA tasks. 

\section{Methods for Learning from Human Feature Importance Supervision}
\label{sec:desiderata_and_methods}

We now describe how to optimize for several key model desiderata using human FI supervision (represented in Fig. \ref{fig:main-fig}). In Sec. \ref{sec:visfis}, we give the overall objective for Visual Feature Importance Supervision (\method), which combines the objective terms below in order to improve model generalization.

\subsection{Accuracy Given Sufficient Information}

\textbf{Goal.} Like \citet{chang2021towards}, we hope for image processing models to make accurate predictions given subsets of image features that are sufficient for arriving at the correct label, since this suggests that a model recognizes that the important features are in fact important.

\vspace{2pt}
\textbf{Method.} 
Access to human explanations should enable us to automatically construct sufficient inputs with some amount of unimportant information removed \cite{chang2021towards}.
In particular, for every input we can create another datapoint by using the human explanation $e$ to \replace unimportant features, while keeping the same label. The corresponding objective is given as:
\vspace{1pt}
\begin{align}
    \mathcal{L}_\textrm{Suff}(\theta, x, y, e) = \textrm{CrossEnt}(f_\theta(x_e), y).
\end{align}
\vspace{1pt}
This objective differs from previous instantiations \cite{chang2021towards} by virtue of the $\mathrm{Replace}$ function used (see Sec. \ref{sec:visfis}).
We compare $\mathcal{L}_\textrm{Suff}$ against an unsupervised baseline using random feature subsets. That is, a random distribution $\mathcal{D}_s$ specifies how likely it is that we \replace a feature:
\vspace{1pt}
\begin{align}
    \mathcal{L}_{\textrm{Suff-Random}}(\theta, x, y, \mathcal{D}_s) = \mathbb{E}_{s \sim \mathcal{D}_s} \textrm{CrossEnt}(f_\theta(x_s), y) 
    \label{eq:unsupervised_sufficiency}
\end{align}
\vspace{1pt}
which, when training via SGD, is estimated using one sample per datapoint per batch.

\subsection{Uncertainty Given Only Unimportant Information}

\textbf{Goal.} We would prefer for a model to give uncertain outputs for inputs with no important features, meaning the model should give a near-uniform distribution over classes. Since there is no evidence for any given class, the model should not be confident the input belongs in a particular class.

\looseness=-1
\vspace{2pt}
\textbf{Method.} With this goal in mind, \citet{chang2021towards} train models to give less confident outputs for images with important foreground features removed. More specifically, they encourage the model to predict any class \emph{except} the image's true class. 
In contrast, we penalize a KL divergence between the model output distribution and a uniform distribution,
\vspace{1pt}
\begin{align}
    \mathcal{L}_{\textrm{Unc}}(\theta, x, e) = \textrm{KL}\big(\textrm{Unif}(|\mathcal{Y}|), f_\theta(x_{u})\big)
\end{align}
\vspace{1pt}
where $\textrm{Unif}(|\mathcal{Y}|)$ is the uniform distribution and $u=1-e$ indicates unimportant features.

\subsection{Invariance to Unimportant Information}
\label{sec:desiderata_inv}

\looseness=-1
\textbf{Goal.} 
We would like models to be invariant to changes in an image's unimportant features. 
This property is desirable because it means that a model correctly treats unimportant features as unimportant. 

\textbf{Method.} 
We first describe a simple data augmentation approach, then describe an FI supervision approach similar to past work \cite{simpson2019gradmask, chang2021towards}.

\looseness=-1
In a data augmentation approach, we train a model to produce the same outputs for two inputs that share the same important information while differing in the unimportant information they contain. 
Specifically, we use $e$ to obtain an input with both important and unimportant features, denoted by $x_{e \cup u} = \mathrm{Replace}(x, e \cup u)$. Then, we penalize the KL divergence between the output distributions on the two inputs $x_e$ and $x_{e\cup u}$.
The resulting objective is then:
\vspace{2pt}
\begin{align}
  \mathcal{L}_{\textrm{Inv-DA}}(\theta, x, e, \mathcal{D}_u) = \mathbb{E}_{u \sim \mathcal{D}_u} \hspace{1pt} \textrm{KL}(f_\theta(x_e), f_\theta(x_{e\cup u}))
  \label{eq:inv_to_unimportant}
\end{align}
\vspace{2pt}
where $D_u$ is a distribution over binary vectors. $\mathcal{L}_\textrm{Inv-DA}$ is estimated with one sample like $\mathcal{L}_\textrm{Suff-Random}$. 

In an FI supervision approach, we first obtain model explanations at the datapoint level as $\tilde{e} = \mathrm{Explain}(f_\theta, x, \hat{y})$, where $\mathrm{Explain}$ is a differentiable explanation method (possible methods described below). Then we seek to directly penalize models for treating unimportant features as important.
To do so, we encourage FI scores for unimportant features to be $0$:
\vspace{2pt}
\begin{align}
    \mathcal{L}_{\textrm{Inv-FI}}(\tilde{e}_u) = ||\tilde{e}_u||_1
    \label{eq:unimportant_to_zero}
\end{align}
\vspace{2pt}
where $\tilde{e}_u$ is the subset of the explanation over features marked as unimportant by $e$. Past work uses an $\ell_2$ distance for this objective \cite{simpson2019gradmask, chang2021towards}, while we use an $\ell_1$ penalty after normalizing explanations to unit length, since explanations from different FI methods have different scales.

\looseness=-1
We consider a few options for differentiable explanation methods.
Past work has primarily used gradient-based
\cite{ross2017right, selvaraju2019taking, simpson2019gradmask, ghaeini2019saliency, wu2019self, chang2021towards} and attention-based explanations \cite{zhong2019fine, stacey2022supervising, gao2022aligning, chrysostomou2021enjoy}. We adopt existing gradient/attention methods and provide new differentiable perturbation-based methods. 
\begin{enumerate}[itemsep=2pt, wide=0pt, leftmargin=*, after=\strut]
\vspace{-1pt}
\item \emph{Gradient-based explanations.} One can optimize objectives involving gradient-based explanations w.r.t. $\theta$ by computing second derivatives like
$\nabla_\theta \nabla_x f_\theta(x)$ in a framework like PyTorch \cite{paszke2017automatic}. We use a simple Vanilla Gradient method and the Expected Gradients method (see Appendix \ref{app:explanation_methods}).
\item \emph{Attention-based explanations.} 
We supervise bounding box attention weights in the UpDn model \cite{anderson2018bottom}, but early experiments suggest this is not an effective method and we do not explore it further.
\looseness=-1
\item \emph{Perturbation-based explanations.} Perturbation-based methods like SHAP \cite{lundberg_unified_2017} are very popular explanation methods, but have seen only limited use for FI supervision \cite{kennedy2020contextualizing}. 
We consider a leave-one-out method (LOO), a keep-one-in method (KOI), Average Effect, and SHAP (see Appendix \ref{app:explanation_methods}).
\end{enumerate}

In Appendix \ref{app:FI_budgets}, we discuss limits on the compute budget used for each method during model training. 

\subsection{Aligned Model and Human Feature Importance}

\textbf{Goal.} 
Alignment between human and model explanations has frequently been proposed as a goal for models
\cite{selvaraju2019taking, shrestha-etal-2020-negative, yang2020object, friedrich2022typology}. 
In general, past works assume that model explanations are \emph{faithful}, meaning they accurately communicate a model's internal reasoning \cite{jacovi2020towards}. 
This assumption is necessary for the alignment between model and human explanations, termed \emph{plausibility} by \citet{jacovi2020towards}, to be evidence that model reasoning is similar to human reasoning. 
Of course, model explanations are not guaranteed to be faithful.
To the extent that they are faithful, however, encouraging explanation plausibility during training may help align model reasoning with human reasoning. 

\vspace{2pt}
\textbf{Method.} 
We first obtain model explanations at the datapoint level as $\tilde{e} = \mathrm{Explain}(f_\theta, x, \hat{y})$ (see Sec. \ref{sec:desiderata_inv} above). 
Then, we can measure the difference between $\tilde{e}$ and the human explanation $e$ using an $l_p$ distance, cosine similarity, or a differentiable ranking function \cite{selvaraju2019taking, wu2019self}.
We use a cosine similarity since model explanations and human explanations do not share the same scale. Our objective is thus:
\vspace{2pt}
\begin{align}
    \mathcal{L}_{\textrm{align}}(\theta, x, e, \tilde{e}) = \textrm{cos-sim}(e, \tilde{e})
    \label{eq:explanation_alignment}
\end{align}

\subsection{Overall Objective for \method: Visual Feature Importance Supervision}
\label{sec:visfis}

\looseness=-1
We combine the supervised objective terms from above to achieve the corresponding model desiderata simultaneously. 
Following objective tuning experiments showing that Inv-FI outperforms Inv-DA (see Appendix Table \ref{tab:full_objective_ablation_table}), we use Inv-FI rather than Inv-DA, and therefore our final \method objective is:
\vspace{2pt}
\begin{align}
    \lambda_1\mathcal{L}_\textrm{Task} + \lambda_2\mathcal{L}_\textrm{Suff} + \lambda_3\mathcal{L}_\textrm{Unc} + \lambda_4\mathcal{L}_\textrm{Align} + \lambda_5\mathcal{L}_\textrm{Inv-FI}
\end{align}
\vspace{2pt}
where $\mathcal{L}_\textrm{Task}$ is a standard supervised cross-entropy loss.
Besides tuning the values of $\lambda_i$ one at a time, we also tune the \replace function and FI method used in this objective, making sure to use comparable compute budgets across FI methods. \replace functions we consider are listed in Appendix \ref{app:replace_functions} (results in Table \ref{tab:replace_ablation_table}), and FI methods in Appendix \ref{app:explanation_methods} (results in Tables \ref{tab:FI_methods_tuning} and \ref{tab:FI_method_ablation_table}). Following tuning, we find that it is preferable to \replace bounding box representations with the negative ones vector, $\{-1\}^d$, and surprisingly, we find that Vanilla Gradient is the best performing FI method, surpassing all perturbation-based methods as well as the Expected Gradients method.

\section{Metrics} 
\label{sec:metrics}

Next, we describe the RRR and explanation metrics for each of our model desiderata outlined above. 
We also measure model ID and OOD accuracy (distribution shifts described in Sec. \ref{sec:experiment_setup}). As with the model objectives, we use the All-Negative-Ones \replace function as needed.

\vspace{2pt}
\textbf{RRR-Suffiency.} We measure model accuracy on inputs containing only features selected as important by their respective human explanation (similar to \cite{chang2021towards}). The remaining features are \replaced.

\looseness=-1
\vspace{2pt}
\textbf{RRR-Uncertainty.} We propose to measure how uncertain the model prediction is given only unimportant information. Specifically, we report the average model predicted probability when we provide only unimportant features to the model (according to the human explanation), so lower is better. 

\vspace{2pt}
\textbf{RRR-Invariance.} We propose to calculate the agreement between model predictions with the input $x_e$ and three $x_{e\cup u}$ that each include a random number of unimportant features.
The final metric is averaged over three random $u$ for each test point, then over all test points. 

\vspace{2pt}
\textbf{Explanation Plausibility.} Our explanation plausibility metric is the Spearman's rank correlation between the human and model feature importance vectors, similar to past work \cite{das2017human, selvaraju2019taking}.
We use continuous FI estimates in order to calculate the rank correlation. A rank correlation is preferrable here because human and model FI explanations do not lie in the same space.

\vspace{2pt}
\looseness=-1
\textbf{Explanation Faithfulness.} We use two standard faithfulness metrics \cite{deyoung_eraser_2019}. Sufficiency measures whether \emph{keeping} important features (according to model explanation $\tilde{e}$) leads the model to \emph{retain} its confidence in its original prediction: $\textrm{Suff}(f_\theta, x, \tilde{e}) = f_\theta(x)_{\hat{y}} - f_\theta(x_{\tilde{e}})_{\hat{y}}$.
Comprehensiveness measures whether \emph{removing} important features from an input leads to a \emph{decline} in model confidence,
$\textrm{Comp}(f_\theta, x, \tilde{e}) = f_\theta(x)_{\hat{y}} - f_\theta(x_{\tilde{e}}^C)_{\hat{y}}$,
where $x_{\tilde{e}}^C = \mathrm{Replace}(x, 1-\tilde{e})$ is the \emph{complement} of features in $x_{\tilde{e}}$. 
We average these score over several sparsity levels of $\tilde{e}$, keeping or removing the top 10\%, 25\%, or 50\% of features \cite{deyoung_eraser_2019}. 
Note we compute these metrics using the best available explanation method per dataset, as measured by explanation faithfulness (comparison in Appendix \ref{app:additional_results}).

\section{Experiment Setup}
\label{sec:experiment_setup}

\looseness=-1

\textbf{Datasets.} 
We perform experiments on three benchmark datasets: CLEVR-XAI \cite{arras2022clevrxai}, GQA \cite{hudson2018gqa}, and VQA-HAT \cite{das2017human}. 
CLEVR-XAI is an algorithmically generated dataset based on CLEVR \cite{johnson2017clevr} and provides ground truth visual segmentation masks for each question. 
CLEVR-XAI is limited in visual varieties and vocabularies, but it offers FI supervision in a controlled, low-noise setting.
GQA contains compositional reasoning questions over naturalistic images. GQA also includes the program for generating the questions and the ground-truth scene graph from the Visual Genome dataset \cite{krishnavisualgenome}. This allows us to obtain bounding boxes of relevant objects identified through the question program, which we use as FI supervision. 
VQA-HAT is based on VQAv1 \cite{antol2015vqa}, including naturalistic images and questions with mouse tracking de-blurring used to collect image FI annotations from humans. For VQA, we report model performance on the more challenging \emph{other} type questions as recommended by \citet{teney2020value}.

\begin{wraptable}{r}{.51\textwidth}
\begingroup
\setlength{\tabcolsep}{4pt}
  \small
  \begin{center}
  \vspace{-21pt}
  \caption{Dataset split sizes.} 
  \begin{tabular}{l r r r r} 
    \toprule
    Dataset & \multicolumn{1}{c}{Train} & \multicolumn{1}{c}{Dev} & \multicolumn{1}{c}{Test-ID} & \multicolumn{1}{c}{Test-OOD}\\ 
    \midrule
     CLEVR-XAI & 83k & 14k &  21k & 22k \\
     GQA-101k & 101k &  20k &  20k & 20k \\
     VQA-HAT &  36k &  6k & 9k & 9k \\  
  \bottomrule 
  \end{tabular}
  \vspace{-14pt}
  \label{tab:dataset}
  \end{center}
\endgroup
\end{wraptable}

\looseness=-1
\vspace{2pt}
\textbf{Distribution Shifts.} 
We create both ID and OOD test sets for each dataset, so we always have four data splits: Train, Dev, Test-ID, and Test-OOD (split sizes shown in Table \ref{tab:dataset}). 
To obtain OOD data, we use distribution shifts similar to those in VQA-CP, which are intended to vary the linguistic bias between ID and OOD splits \cite{agrawal2018vqacp}. We apply the same procedure for distribution shift on all three datasets for comparability.
In detail, we create groups of questions according to the first few words in each question (indicating the type of question), and allocate groups unevenly into ID and OOD sets, randomly assigning 80\% of each group to one set and 20\% to the other.
The ID set is split into Train, Dev, and Test-ID. Model selection is done according to Dev set performance. We further downsample the very large GQA dataset from to about 100k for training and 20k for other splits. See Appendix Fig. \ref{fig:training_size_ablation} for training size ablation analysis. We note that we avoid several pitfalls in evaluating VQA models against distribution shifts, as outlined by \citet{teney2020value}. See Appendix Table \ref{tab:resplit_sensitivity} for sensitivity analysis with randomly resplit data.

\vspace{2pt}
\textbf{Human Feature Importance.} 
For all datasets, we obtain human FI scores at the bounding box (BB) level for detected BBs from the Faster-RCNN detector \cite{ren2015faster-rcnn}.
Following \citet{selvaraju2019taking}, for both VQA-HAT and CLEVR-XAI we obtain importance scores from pixel-level annotations as $s^k = E_i^k / (E_i^k + E_o^k)$, where $s^k$ is the score for the $k$\textsuperscript{th} detected BB and $E_i^k$ and $E_o^k$ are the average pixel-level importance score inside and outside the BB, respectively. 
VQA-HAT has real-valued pixel-level scores, while for CLEVR-XAI, we set the pixel-level score to 1 for pixels within the segmentation mask and 0 elsewhere.
For GQA, since we have BB level annotations, we calculate the importance score based on the intersection over union (IoU) between ground-truth important BBs and detected BBs:
$s^k = \max_{l\in \mathcal{G}} \textrm{IoU}(B_d^k, B_{\textrm{gt}}^l)$
where $B_d^k$ is the BB of the $k$\textsuperscript{th} detected object and $B_{\textrm{gt}}^l$ is the BB of the $l$-th ground-truth object. With importance scores for each BB, we manually set a threshold for determining important and unimportant objects (0.85, 0.55, and 0.3 for CLEVR-XAI, VQA-HAT, and GQA respectively).
See Appendix \ref{app:data_details} for sensitivity analysis for this threshold. 

\vspace{2pt}
\looseness=-1
\textbf{Models.} We run experiments with UpDn \cite{anderson2018bottom} and LXMERT \cite{tan2019lxmert}. 
Both models rely on bounding box representations generated by a pretrained Faster R-CNN model \cite{ren2015faster-rcnn} (further details in Appendix \ref{app:training_details}).

\vspace{2pt}
\textbf{Hypothesis Testing.} We conduct hypothesis tests via a bootstrap resampling model seeds and datapoints 10k times \cite{efron1994introduction}. We obtain 95\% confidence intervals in the same way. 

\section{Experiment Results}
\label{sec:results}

\subsection{Can FI Supervision Improve Model Accuracy for VQA?}
\label{sec:supervision_improve_accuracy}

\looseness=-1
\textbf{Design.} 
Using UpDn on our three datasets, we compare \method with previous state-of-the-art FI supervision methods for VQA tasks \cite{selvaraju2019taking, wu2019self} as well as for image classification \cite{simpson2019gradmask, chang2021towards}.
We give results for LXMERT only on CLEVR-XAI, since GQA and VQA are a part of the pretraining data for LXMERT \cite{tan2019lxmert}. Note we test on the more challenging \emph{other} type questions only for VQA, following \citet{teney2020value}.
\citet{selvaraju2019taking} use $\mathcal{L}_\textrm{align}$ with a ranking loss to align Vanilla Gradient explanations and human FI supervision. \citet{wu2019self} propose a relaxed version of the ranking loss that binarizes important and unimportant features according to human FI supervision and encourages higher model FI for important objects than unimportant ones.
The other methods we consider all use an $\mathcal{L}_\textrm{FI-Inv}$ objective with an $l_2$ penalty on Vanilla Gradient explanations. On top of this, \citet{chang2021towards} add an $\mathcal{L}_\textrm{Suff}$ objective with a Shuffle $\mathrm{Replace}$ function that randomly permutes features rather than replacing them, to preserve the marginal data distribution, and \citet{singla2022core} add an $\mathcal{L}_\textrm{Suff}$ objective with a Gaussian noise $\mathrm{Replace}$ function.
Our unsupervised baselines are models trained with only label supervision or using $\mathcal{L}_\textrm{Suff-Random}$. 

\begingroup
\setlength{\tabcolsep}{5pt}
\begin{table}[h!]
\small
\begin{center}
\vspace{-3pt}
\caption{Test accuracy across FI supervision methods and datasets with an UpDn model. 
We bold/underline numbers higher than the best unsupervised baseline at a significance threshold of $p<.05$ (and bold is better than underline at $p<.05$).
}
\vspace{-3pt}
\begin{tabular}{l r r r r r r}
\toprule
& \multicolumn{2}{c}{CLEVR-XAI} & \multicolumn{2}{c}{GQA-101k} & \multicolumn{2}{c}{VQA-HAT} \\
\cmidrule(lr){2-3} \cmidrule(lr){4-5}  \cmidrule(lr){6-7}
Method & \multicolumn{1}{c}{ID} & \multicolumn{1}{c}{OOD} & \multicolumn{1}{c}{ID} & \multicolumn{1}{c}{OOD} & \multicolumn{1}{c}{ID} & \multicolumn{1}{c}{OOD} \\
    \midrule
    Baseline & 71.37±0.57 & 36.80±1.00 & 51.82±0.62 & 31.80±0.64 & 37.53±1.32 & 28.76±1.10   \\
    Suff-Random & 71.72±0.57 & 39.08±0.80 & 51.59±0.65 & 31.65±0.82 & 37.99±1.35 & 29.34±1.03 \\
    \citet{selvaraju2019taking} & 71.32±0.58 & 37.96±1.00  & 51.38±0.62 & 31.99±0.77 & 36.93±1.37 & 27.38±1.27  \\
    \citet{wu2019self} & 71.48±0.64 & 37.31±0.86  & 51.54±0.67 & 31.61±0.78 & 37.24±1.32 & 28.26±1.15   \\
    \citet{simpson2019gradmask} & 71.22±0.60 & 37.54±0.71 & 52.10±0.68 & 31.99±0.77 & 37.66±1.30 & 28.73±1.44    \\
    \citet{chang2021towards} & 70.77±0.56 & 35.38±0.92 & 50.29±0.65 & 30.40±0.86 & 32.55±1.41 & 17.98±1.75   \\
    \citet{singla2022core} & 71.54±0.58	& 38.25±1.39 & \underline{52.42}±0.66 & \underline{32.58}±0.59 & 38.28±1.37 & 29.25±2.12 \\
    \method & \textbf{72.82}±0.56 & \textbf{43.78}±1.11 & \textbf{54.81}±0.61 & \textbf{34.88}±0.80 & \textbf{38.75}±1.35 & \textbf{31.21}±1.28  \\
    \midrule 
    \ w/ Rand. Supervis. & 69.70±0.67 & 33.28±1.03 & 49.82±0.62 & 29.93±0.89 & 37.16±1.30 & 27.51±1.17  \\
    \bottomrule
\end{tabular}
\label{tab:main_table}
\end{center}
\vspace{-15pt}
\end{table}
\endgroup

\begin{wraptable}{r}{.48\textwidth}
\begingroup
\setlength{\tabcolsep}{5pt}
\small
\begin{center}
\vspace{-13pt}
\caption{LXMERT + CLEVR-XAI results.}
\vspace{3pt}
\begin{tabular}{l r r}
\toprule
Method & \multicolumn{1}{c}{ID Acc} & \multicolumn{1}{c}{OOD Acc} \\
    \midrule
    Baseline & 86.91±0.43 & 73.76±0.72   \\
    Suff-Random & 86.53±0.47 & 73.52±1.07 \\
     \citet{selvaraju2019taking} & 87.03±0.43 & 74.56±0.58  \\
    \citet{wu2019self} & 86.73±0.46 & 73.97±0.75 \\
    \citet{simpson2019gradmask} & 86.22±0.57 & 73.12±1.28 \\
    \citet{chang2021towards} & 85.05±0.57 & 67.27±2.27  \\
    \citet{singla2022core} & 87.08±0.46 & 74.10±0.75  \\
    \method & \textbf{87.39}±0.45 & \textbf{74.83}±0.70 \\
    \midrule
    \ w/ Rand. Supervis.  & 85.84±0.83 & 71.81±1.34  \\
    \bottomrule
\end{tabular}
\label{tab:lxmert_main}
\end{center}
\vspace{-9pt}
\endgroup
\end{wraptable}

\looseness=-1
\textbf{Results.} We show results for UpDn in Table \ref{tab:main_table} and for LXMERT in Table \ref{tab:lxmert_main}. 
First, we find that FI supervision can meaningfully improve model accuracy. 
With UpDn on CLEVR-XAI, \method improves ID accuracy by 1.1 points (±0.5; $p{=}$1e-4) and OOD accuracy by 4.7 points (±1.4; $p{<}$1e-4) over the strongest baseline without supervision, Suff-Random (see Appendix Fig. \ref{fig:xai_barplot} for breakdown in improvements by CLEVR question type). 
Trends are similarly positive on the other datasets and with LXMERT, where \method outperforms the baseline by 0.48 points (±0.35; $p{<}$.01) on ID data and 1.07 points ($p{<}$1e-4) on OOD data (for results on all VQA question types, see Appendix Table \ref{tab:full_vqa_table}). 
These improvements do not persist when using random explanations (last row), meaning they are caused by the human supervision. 
Finally, we observe that \method is the best overall method across datasets and architectures, as other methods typically do not improve accuracy over an unsupervised baseline. The next best method is that of \citet{singla2022core}, which improves over an unsupervised baseline only for the GQA dataset with UpDn, but \method still outperforms \citet{singla2022core} there by 2.39 (±0.55; $p{<}$1e-4) points on ID data and 2.31 points (±0.66; $p{<}$1e-4) on OOD data.

\begin{wrapfigure}[22]{r}{0.5\textwidth}
  \vspace{-21pt}
  \begin{center}
    \includegraphics[width=.5\textwidth]{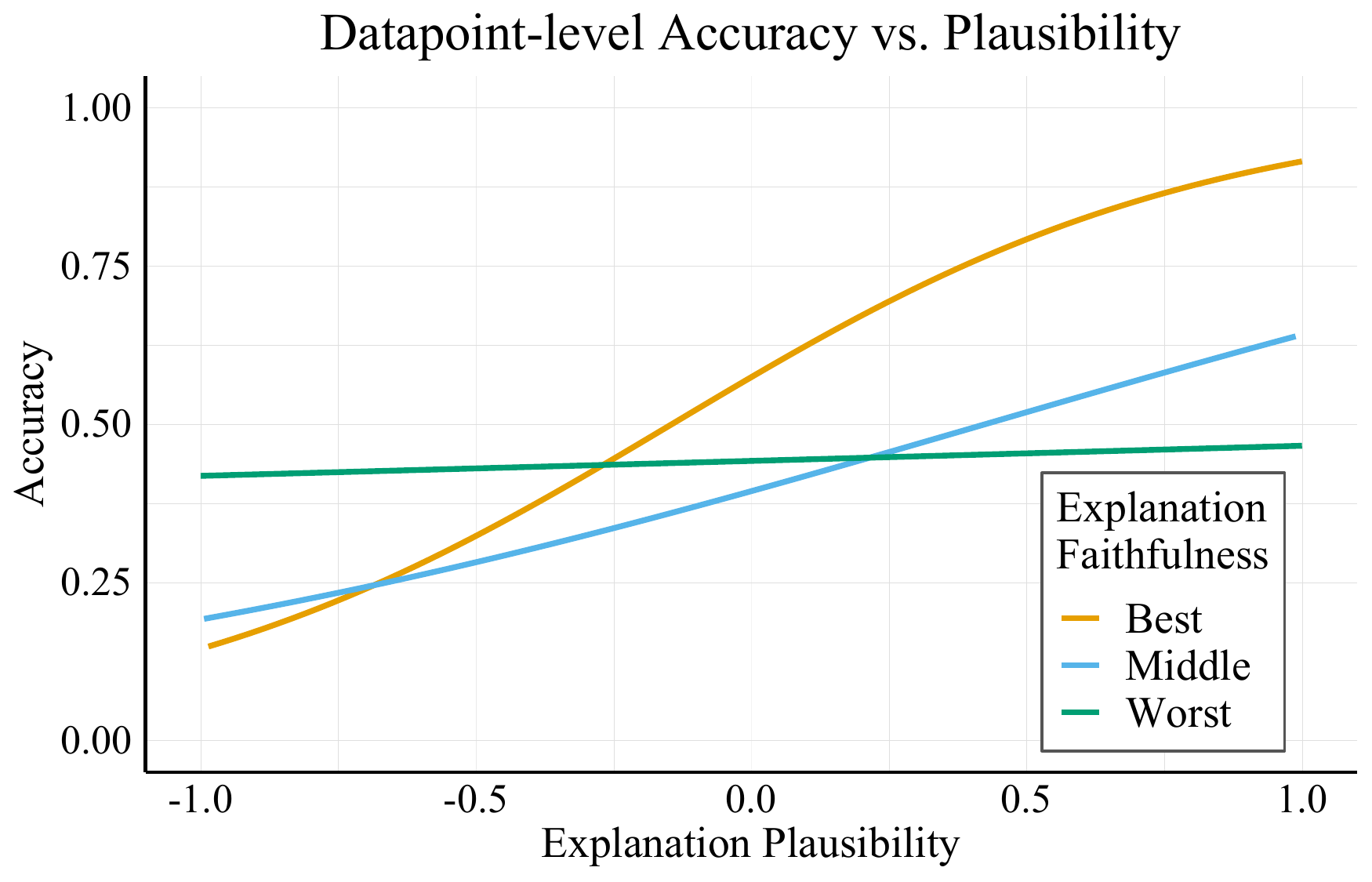}
  \end{center}
  \vspace{-11pt}
  \caption{
  Datapoint-level accuracy by explanation plausibility, averaged across CLEVR-XAI models. Trendlines are logistic regressions.
  When explanations are more faithful, their alignment with human explanations better correlates with model accuracy.
  }
  \label{fig:acc_by_plau}
  \vspace{-2pt}
\end{wrapfigure}

\subsection{How Does FI Supervision Improve Accuracy?} 

\textbf{Design.} Past work hypothesizes that FI supervision improves accuracy by aligning model and human FI \cite{selvaraju2019taking, chang2021towards}. Surprisingly, we find that the relationship between model test accuracy and average explanation plausibility is fairly weak (linear correlation on UpDn+CLEVR-XAI models is $\rho{=}$0.14±0.19).
Here, we argue that plausible explanations alone are not evidence of correct model reasoning, but plausible \emph{and} faithful explanations are. 
Using 4 million ID/OOD test predictions from UpDn+CLEVR-XAI models, we visualize trendlines from logistic regressions predicting model accuracy based on plausibility and faithfulness at the datapoint level, grouped into Worst, Middle, and Best faithfulness categories based on Sufficiency/Comprehensiveness metrics (see Appendix \ref{app:data_details}). 

\begin{figure}[t]
  \vspace{-4pt}
  \begin{center}
    \includegraphics[width=0.98\textwidth]{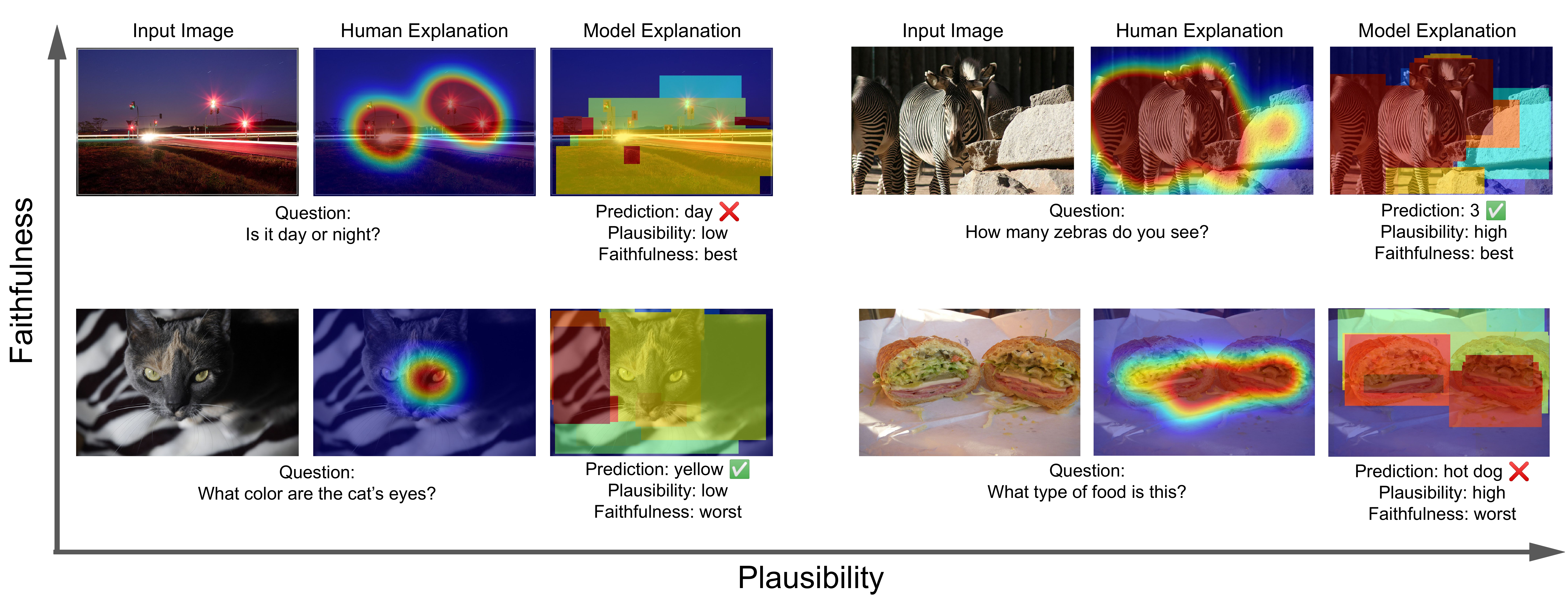}
  \end{center}
  \vspace{-11pt}
  \caption{Qualitative visualization of the relationship between accuracy, plausibility, and faithfulness represented in Fig. \ref{fig:acc_by_plau}.
  In a low faithfulness setting (in terms of explanation sufficiency), a data point with an implausible explanation can still have a correct prediction (bottom left), while a data point with highly plausible explanation can have an incorrect prediction (bottom right). Among higher faithfulness points (top row), data with more plausible explanations tend to be correctly predicted.}
  \label{fig:visualization}
  \vspace{-11pt}
\end{figure}

\looseness=-1
\textbf{Results.} Fig. \ref{fig:acc_by_plau} shows that as an explanation for a datapoint becomes more plausible, the model is more likely to correctly predict that point's label, but \emph{only when the explanation is also faithful}. 
Indeed, a maximally plausible and faithful explanation has about a 90\% chance of being correct, while a minimally plausible but highly faithful explanation has closer to a 12.5\% of being correct. For unfaithful explanations, plausibility has essentially no relationship with accuracy.
Though these trends are not necessarily causal, they are consistent with the view that when model predictions are correct, it is because their true reasoning (as revealed by \emph{faithful} explanations) aligns with human reasoning. 
Fig. \ref{fig:visualization} qualitatively illustrates this relationship among faithfulness, plausibility, and accuracy with example data points and model predictions.
We emphasize that while past work has treated plausibility as an RRR metric \cite{selvaraju2019taking, shrestha-etal-2020-negative, yang2020object, prasad2020extent, friedrich2022typology}, the results here demonstrate that plausibility alone cannot be a measure of model correctness.
\vspace{-2pt}

\subsection{Which FI Supervision Objectives Improve Accuracy?}
\label{sec:supervision_objective_ablation}

\textbf{Design.} We ablate across objective terms from Sec. \ref{sec:desiderata_and_methods} for UpDn on CLEVR-XAI. 
The weight for each objective term is tuned while using only that objective, then kept fixed when objectives are combined (further details in Appendix \ref{app:training_details}). We consider another kind of ablation experiment where we use random supervision for one objective at a time in \method, with results in Appendix Table \ref{tab:random_sup}.

\vspace{2pt}
\textbf{Results.} In Table \ref{tab:main_objective_ablation_table}, we find that
each individual objective is valuable on its own, and they do well when combined.
Relative to the Baseline OOD accuracy, Suff-Human adds 4.1 points, Unc adds 1.54 points, Inv-FI adds 2.08 points, and Align adds 4.81 points. When the four objectives are combined in \method, the improvement rises to 6.98 points. 

\vspace{-1pt}
\subsection{Can FI Supervision Make Models Right for the Right Reasons?}
\label{sec:supervision_make_models_RRR}

\begin{wraptable}{r}{.61\textwidth}
\begingroup
\setlength{\tabcolsep}{5pt}
\small
\begin{center}
\vspace{-23pt}
\caption{Objective ablation for UpDn + CLEVR-XAI}
\vspace{1pt}
\begin{tabular}{l r r r r r r}
\toprule
& \multicolumn{2}{c}{Acc} & \multicolumn{3}{c}{RRR Metrics} & \multicolumn{1}{c}{Expl.} \\
\cmidrule(lr){2-3} \cmidrule(lr){4-6}  \cmidrule(lr){7-7}
Objective & \multicolumn{1}{c}{ID } & \multicolumn{1}{c}{OOD } & \multicolumn{1}{c}{Suff } & \multicolumn{1}{c}{Inv } & \multicolumn{1}{c}{Unc} & \multicolumn{1}{c}{Plaus.} \\
    \midrule
    Baseline & 71.37 & 36.80 & 48.82 & 77.89 & 55.17 & 28.82 \\
    Inv-DA & 71.17 & 35.91 & 72.53 & \textbf{93.12} & 76.29 & 14.33 \\
    Inv-FI & 71.41 & 38.88 & 45.31 & 76.34 & 71.41 & 28.60  \\
    Unc & 71.30 & 38.34 & 10.75 & 86.58 & \textbf{4.16} & 8.56 \\
    Align & 72.04 & 41.61 & 61.19 & 79.51 & 64.22 & \textbf{37.20} \\
    Suff-Random & 71.73 & 39.08 & 73.59 & 92.59 & 60.93 & 17.32 \\
    Suff-Human & 71.87 & 40.91 & 76.94 & 90.82 & 81.42 & 16.27 \\
    + Align & 72.42 & 41.63 & 78.55 & 89.69 & 80.02 & 35.73 \\
    + Unc & 72.33 & 41.54 & \textbf{77.83} & 89.70 & 41.68 & 23.41 \\
    \method & \textbf{72.82} & \textbf{43.78} & 76.65 & 91.72 & 43.64 & 22.67 \\
    \bottomrule
\end{tabular}
\label{tab:main_objective_ablation_table}
\end{center}
\vspace{-19pt}
\endgroup
\end{wraptable}

\textbf{Design.} We report RRR metrics as well as explanation plausibility for the UpDn+CLEVR-XAI models from our objective ablation above. 

\looseness=-1
\textbf{Results.} In Table \ref{tab:main_objective_ablation_table}, we find that FI supervision generally improves RRR metric scores. 
Compared to the Baseline, \method achieves 27.8 points better Sufficiency, 13.8 points better Invariance, and 11.5 points better Uncertainty. 
The best unsupervised method closes the gap slightly on RRR-Suff and RRR-Inv. Specifically, Suff-Random is only 3.97 points worse than Suff-Human on RRR-Suff, and only 0.53 points worse than Inv-DA on RRR-Inv. It suggests that FI supervision noticeably improves RRR metrics, meaning model behavior better fulfills the theoretical desiderata from Sec. \ref{sec:desiderata_and_methods}. 

\vspace{-2pt}

\subsection{Do RRR Metrics Predict OOD Generalization?}
\label{sec:predict_OOD_generalization}

\vspace{-1pt}

\looseness=-1
\textbf{Design.} 
We measure the correlation between RRR metrics (calculated with ID data) and OOD accuracy across a large set of models. We report results here for all UpDn models on CLEVR-XAI, with similar results for GQA/VQA and LXMERT given in Appendix Table \ref{tab:metric_table_others}.
We consider a few possible model metrics, including several composite metrics that combine model-level metrics.
To optimally weight the individual metrics, we fit statistical models to the data that predict OOD accuracy given the available metrics. Since this risks overfitting the composite metrics to the data we have, we perform a cross-validation resampling model-level statistics 10k times, using 90 models' metrics as training data and 10 for testing each time. The final metrics we consider are: (1) ID accuracy on its own as a baseline, (2) RRR metrics on their own, (3) ID accuracy plus average model confidence, (4) ID accuracy plus explanation metrics (for plausibility and faithfulness), (5) ID accuracy plus RRR metrics, and (6) All Metrics, which uses all available metrics.

\begin{wraptable}{r}{0.46\textwidth}
\small
\vspace{-23pt}
\begin{center}
\caption{Correlations between metrics and OOD accuracy, with 95\% confidence intervals.}
\vspace{1pt}
\begin{tabular}{l r r}
\toprule
& \multicolumn{2}{c}{$\rho$\hspace{1pt}(\textrm{Metric}, \textrm{OOD Acc})} \\
\cmidrule(lr){2-3}
Metric & \multicolumn{1}{c}{Train} & \multicolumn{1}{c}{Test} \\
    \midrule
    RRR-Suff & 0.278 & 0.333 ±.0021 \\
    RRR-Inv & 0.149 & 0.157 ±.0058 \\
    RRR-Unc & 0.029 & 0.021 ±.0063 \\
    ID Acc & 0.870 & 0.863 ±.0018 \\
    + Model Conf. & 0.909 & \textbf{0.907} ±.0010 \\
    + Expl. Metrics & 0.875 & 0.861 ±.0046 \\
    + RRR-all & 0.874 & 0.852 ±.0033 \\
    All Metrics & \textbf{0.925} & 0.891 ±.0014  \\
    \bottomrule
\end{tabular}
\label{tab:metric_table}
\end{center}
\vspace{-8pt}
\end{wraptable}

\textbf{Results.} In Table \ref{tab:metric_table}, we show the average correlations between each metric and model OOD accuracy achieved in our cross-validation. 
Interestingly, we find that \textbf{RRR metrics do not achieve a better correlation with OOD accuracy than ID accuracy does on its own}. 
ID accuracy alone has a correlation of 0.863 with OOD accuracy, while using ID Acc + RRR metrics achieves a correlation of 0.852.
In fact, the only additional metric that improves one's ability to predict OOD accuracy is the average model confidence on ID data (more confident models have slightly better OOD accuracy), though this does not hold for LXMERT models (see Appendix Table \ref{tab:metric_table_others}). 
These results cast doubt on the value of RRR metrics. 
If ID accuracy on its own is a better predictor of OOD accuracy than RRR metrics, then RRR metrics may not be a better measure of the quality of model reasoning than ID accuracy is. 
We believe the RRR metrics considered in this paper are still theoretically justified as model desiderata, but we cannot recommend them as measures of model generalization to OOD data.

\vspace{-2pt}
\section{Discussion \& Conclusion}
\label{sec:discussion}
\vspace{-1pt}
\textbf{Limitations.} Though we evaluate with two standard model architectures and three datasets, our conclusions may be limited to settings using Faster-RCNN \cite{ren2015faster-rcnn} bounding box representations as the feature space rather than pixel space. 
Additionally, though we follow existing guidelines with our distribution shifts \cite{das2017human, teney2020value}, we do not measure model generalization across all typical kinds of shifts \cite{quinonero2008dataset}. 
Lastly, we note that FI supervision methods are limited by the need for additional annotations.

\looseness=-1
\textbf{Ethics.} 
We hope that our findings regarding model accuracy and explanation plausibility/faithfulness will help dispel the notion that models reason like humans (or are more grounded) simply because model explanations look similar to human explanations, which can cause unwarranted trust in ML models \cite{jacovi2021formalizing}.
We do not foresee specific ethical risks arising from this work that do not already apply to the general use of machine learning for visual question answering tasks, such as the potential deployment of ML models in settings where they may harm people \cite{weidinger2021ethical, suresh2021framework}.

\looseness=-1
\textbf{Conclusions.} In this paper, we show that 
(1) FI supervision can improve VQA model accuracy via our \method method, 
(2) accuracy improvements appear to stem from improving explanation plausibility specifically for faithfully explained data,
(3) FI supervision can improve RRR metric performance, and 
(4) RRR metrics do not actually correlate well with OOD accuracy. 

\vspace{-1pt}
\looseness=-1
\section*{Acknowledgements} 
\vspace{-1pt}
We thank Jaemin Cho for helpful discussion of this work, as well as Derek Tam, Xiang Zhou, and Archiki Prasad for useful feedback. This work was supported by ARO Award W911NF2110220, DARPA Machine-Commonsense (MCS) Grant N66001-19-2-4031, NSF-AI Engage Institute DRL-211263, and a Google PhD Fellowship. The views contained in this article are those of the authors and not of the funding agency.

\bibliographystyle{plainnat}
\bibliography{main}

\appendix

\section{Feature Importance Explanation Methods}
\label{app:explanation_methods}

We briefly review several FI explanation methods and explain how they are used in this paper. These methods can be classified as gradient-based (1-2), attention-based (3), and perturbation-based (4-7). Note that when computing derivatives of model outputs for explanation methods, we use the logit of the predicted class rather than the predicted probability for purposes of numerical stability.

\begin{enumerate}[itemsep=1pt, wide=0pt, leftmargin=*, after=\strut]
    \item Vanilla Gradient (VGrad) \cite{simonyan_2013_deep}. This method offers an explanation of model behavior in terms of the gradient of the model output with respect to the input, $\nabla_x f_\theta(x)_{\hat{y}}$. When computing scores for a bounding box vector representation, we sum up the gradient for each element.
    \item Expected Gradients (ExpGrad) \cite{erion2021improving}. This method estimates the integral in Integrated Gradients \cite{sundararajan_2017_axiomatic} by Monte Carlo sampling in order to speed up computation, and it uses the data distribution to obtain baseline inputs.
    That is, the explanation
    \begin{align*}
        \tilde{e} = \mathbb{E}_{\alpha \sim \textrm{Unif}(0,1)} \mathbb{E}_{x' \sim D} \bigg[(x' - x) \circ \nabla_x f_\theta\big(x' + \alpha(x-x')\big)_{\hat{y}} \bigg]
    \end{align*}
    is estimated with a single sample of $\alpha$ and $x' \sim D$ using the training dataset $D$. We consider alternative baselines $x'$ later.
    \item Attention weights (AttWeight) \cite{jain2019attention,wiegreffe2019attention, stacey2022supervising}. This approach treats attention weights in a model as an explanation of model feature importance. For the Up-Down model \cite{anderson2018bottom}, we use its sole set of top-down attention weights, but early experiments suggest this is not an effective method and we do not explore it further.
    \item Leave-one-out omission (LOO) \cite{li2016understanding}. An LOO explanation assigns a score to feature $j$ as the difference in the function output on the original input and an input with feature $j$ replaced. Any $\mathrm{Replace}$ function may be used with LOO. Hence $\tilde{e}_j = \mathrm{diff}\big(f(x), \ \mathrm{Replace}(x, \vec{1}_{-j})\big)$ where $\mathrm{diff}$ measures the difference in function outputs, and $\vec{1}_{-j}$ is the ones vector with element $j$ set to 0.
    \item Keep-one-in omission (KOI). The complement of leave-one-out, this method scores each feature by computing the effect of replacing all features except that a given feature.
    \item SHAP \cite{lundberg_unified_2017}. We use the model-agnostic Kernel SHAP method, a generalization of LIME which assigns scores to features by fitting a linear model on perturbations of an input in order to predict the effect of each feature perturbation on the model output. 
    Specifically, Kernel SHAP obtains an explanation by solving a weighted regression problem where model outputs are predicted based on the presence of features in the input:
    \begin{align}
        \argmin_{\tilde{e}} \mathbb{E}_{s \sim \mathcal{D}_s} \ \pi(s) \big(f_\theta(x_s)_{\hat{y}} - f_\theta(x_{\vec{0}})_{\hat{y}} - \tilde{e}^Ts\big)^2
        \label{eq:shap_obj}
    \end{align}
    where $s$ is a random binary mask over features, $x_s=\mathrm{Replace}(x, s)$, the ``null'' input $x_{\vec{0}} = \mathrm{Replace}(x, \vec{0})$, and $\pi$ is the Shapley kernel \cite{lundberg_unified_2017}. 
    Any $\mathrm{Replace}$ function may be used with SHAP.
    \item Average Effect (AvgEffect). This method follows SHAP exactly except for the use of a regression. To estimate a feature's importance, we aim to compute the expected difference between model outputs with that feature observed vs. replaced:
    \begin{align}
        \tilde{e}_j = 
        \mathbb{E}_{s_1, s_0 \sim \mathcal{D}_s}
        \mathrm{diff}\big(
        f_\theta(x_{s_1})_{\hat{y}}
        ,\ 
        f_\theta(x_{s_0})_{\hat{y}}
        \big)
    \end{align}
    where $s_1$ and $s_0$ are versions of a random binary vector $s$ where some element has been set to 1 in $s_1$ and 0 in $s_0$. In practice this expectation is estimated via Monte Carlo sampling. As long as elements of $s$ are sampled independently, this method gives the same result as Kernel SHAP when the number of samples is large, but results will differ when the sample size is small.
\end{enumerate}

\section{\replace Functions}
\label{app:replace_functions}
We explain the tuning process for \replace functions in this section. As the \replace function is used in both obtaining model FI and data augmentation, we tune the \replace functions using the Align and Align+Suff-Human objectives with the LOO explanation method, and we select the function with the highest average Dev set performance. 
For the full sequential tuning process across all hyperparameters, see Appendix~\ref{app:training_details}. 
We consider five different \replace functions: All-Zeros, All-Negative-Ones, Gaussian, Marginal Distribution, and Shuffling. The first two functions simply replace features with zeros or negative ones. Gaussian function adds zero-mean Gaussian noise to input features with the standard deviation calculated using all features within the current batch. Marginal Distribution replaces a feature (a bounding box) with a randomly sampled feature (another bounding box) from the current batch. The Shuffle function shuffles elements of the input representation across all bounding boxes that need replacement within one sample (within and across bounding boxes). We find that All-Negative-Ones \replace function has the highest average accuracy on the Dev set, and we use it for all situations where replacement is needed (see Table \ref{tab:replace_ablation_table}). 

\begin{table}[t!]
\small
\begin{center}
\caption{\replace Functions Tuning}
\vspace{-2pt}
\begin{tabular}{l r r}
\toprule
Method & Align & Align+Suff-Human \\
    \midrule
    All-Zeros & 68.41 & 68.55 \\
    All-Negative-Ones & 68.29 & 70.67 \\
    Gaussian & 68.80 & 69.94 \\
    Marginal Dist & 67.54 & 69.25 \\
    Shuffle & 43.10 & 46.10 \\
    \bottomrule
\end{tabular}
\label{tab:replace_ablation_table}
\end{center}
\vspace{-10pt}
\end{table}

\section{Differentiable SHAP}
\label{app:differentiable_shap}

In this section, we show how to differentiate through SHAP explanations while respecting the theoretical properties that SHAP explanations provide. In Appendix \ref{app:FI_budgets}, we discuss how to limit the computational burden of computing perturbation-based explanations during model training.

Kernel SHAP \cite{lundberg_unified_2017} values are obtained via a weighted linear regression as follows: To explain a model $f : \mathcal{X} \rightarrow \mathcal{Y}$, one defines a data distribution $\mathcal{D}_s$ over binary feature masks for randomly replacing features with some reference value (denoted in our paper by the $\mathrm{Replace}$ operation). In SHAP, these reference values are either (1) randomly drawn from the marginal data distribution over that feature, or (2) preset by the user to a fixed value for all features. We choose the second option based on $\mathrm{Replace}$ function tuning. The closed-form solution for SHAP values is then given by a weighted least-squares regression \cite{lundberg_unified_2017}:
\begin{align}
    \tilde{e} = (S^T W S)^{-1} S^T W Y
    \label{eq:closed_form}
\end{align}
\looseness=-1
where the row vector $S_i$ is drawn from $\mathcal{D}_s$, $W$ is a diagonal weight matrix with elements $W_{ii} = \pi(S_i)$, and $Y_i$ is the difference in function outputs on $x_{S_i}$ and the ``null'' input, $f_\theta(x_{S_i}) - f_\theta(x_{\vec{0}})$.
This formulation can also satisfy the additivity constraint that the explanation weights sum to the difference $f_\theta(x) - f_\theta(x_{\vec{0}})$. This is done by adding a ``data point'' $S_i$ that is all ones, with its weight $W_{ii}$ manually set to a large value.
The resulting explanation is differentiable w.r.t. $\theta$ by virtue of being differentiable w.r.t. $Y$.

\section{Varying Compute Budgets in Feature Importance Methods}
\label{app:FI_budgets}

In Table \ref{tab:FI_method_ablation_table}, we show the performance of each FI method for improving CLEVR-XAI dev ID accuracy with UpDn. Surprisingly, we find that accuracy improvements do not increase with a higher compute budget for the FI method. Below, we describe how the compute budget can vary for each method.

Vanilla Grad and Attention have invariable compute budgets, as measured in terms of the number of forward+backward passes. However, the other methods have variable budgets. Expected Gradients depends straightforwardly on the number of sampled $\alpha$ values, since we use the same negative-ones baseline feature value for all points (one forward and one backward pass per sample).

To compute a perturbation-based explanation, we need to compute $f_\theta$ at least once per feature in $x$. This is because we need to measure the effect of replacing each feature separately. With SHAP, we need at least one sample $S_i$ per feature in order for $\tilde{e}$ to be identifiable (equivalently, for $S^T W S$ to be invertible). 
However, a \emph{complete explanation is not needed for every datapoint during training}. 
Instead, we can estimate feature importance for only a subset of features for each datapoint in a given batch. This allows us to greatly limit the computational cost of explanation supervision. 
In fact, we can use \emph{as little as a single sample per data point}.
The strength of SGD-based training in this context is that, over the course of training, a large number of feature importance estimates will be computed and penalized against human explanations.

With our per-explanation compute budget of $k$ model forward+backward passes and an input dimensionality $d$, we allow for $k<d$ by explaining only $k$ features while keeping the other $d-k$ features constant.
With LOO explanations, this simply requires not computing scores for $d-k$ features, which are ignored in the $\mathcal{L}_{\textrm{align}}$ loss. 
With SHAP explanations, we pick $d-k$ features to always set to 0 in our random masks $s$. Then when computing Eq. \ref{eq:closed_form}, we drop those constant feature columns from $S$ to obtain a new $k \times k$ matrix, which ensures that $\tilde{e}$ is identifiable. 

\section{Data Details}
\label{app:data_details}
\paragraph{Dataset License.} We conduct experiments on three datasets: CLEVR-XAI \cite{arras2022clevrxai} under the CC BY-NC-ND 4.0 license, GQA \cite{hudson2018gqa}, and VQA-HAT \cite{das2017human} both under the CC BY 4.0 license. 
\paragraph{Distribution Shift Resplit Sensitivity.} Since we randomly construct ID and OOD splits with our distribution shift, we show the robustness of \method across three resplits of CLEVR-XAI dataset here. In Table \ref{tab:resplit_sensitivity}, we see that \method gives significant performance improvements in all resplits. The absolute OOD accuracies vary across resplits, but the size of the OOD performance improvement between \method and the baseline is generally similar, with between a 2.3 and 3.4 percentage point improvement for each split.

\begin{table}[t!]
\small
\begin{center}
\caption{Resplit Sensitivity Test.}
\vspace{-2pt}
\begin{tabular}{l r r r r r r}
\toprule
& \multicolumn{2}{c}{Resplit 1} & \multicolumn{2}{c}{Resplit 2} & \multicolumn{2}{c}{Resplit 3} \\
\cmidrule(lr){2-3} \cmidrule(lr){4-5}  \cmidrule(lr){6-7}
FI Method & \multicolumn{1}{c}{ID Acc.} & \multicolumn{1}{c}{OOD Acc.} & \multicolumn{1}{c}{ID Acc.} & \multicolumn{1}{c}{OOD Acc.} & \multicolumn{1}{c}{ID Acc.} & \multicolumn{1}{c}{OOD Acc.} \\
    \midrule
    Baseline & 69.99 & 50.40 & 69.21 & 57.73 & 69.08 & 58.75 \\
    \method & 72.00 & 52.79 & 71.03 & 60.28 & 70.70 & 62.14 \\
    \bottomrule
\end{tabular}
\label{tab:resplit_sensitivity}
\end{center}
\vspace{-4pt}
\end{table}

\begin{figure}
  \vspace{-8pt}
  \begin{center}
    \includegraphics[width=.69\textwidth]{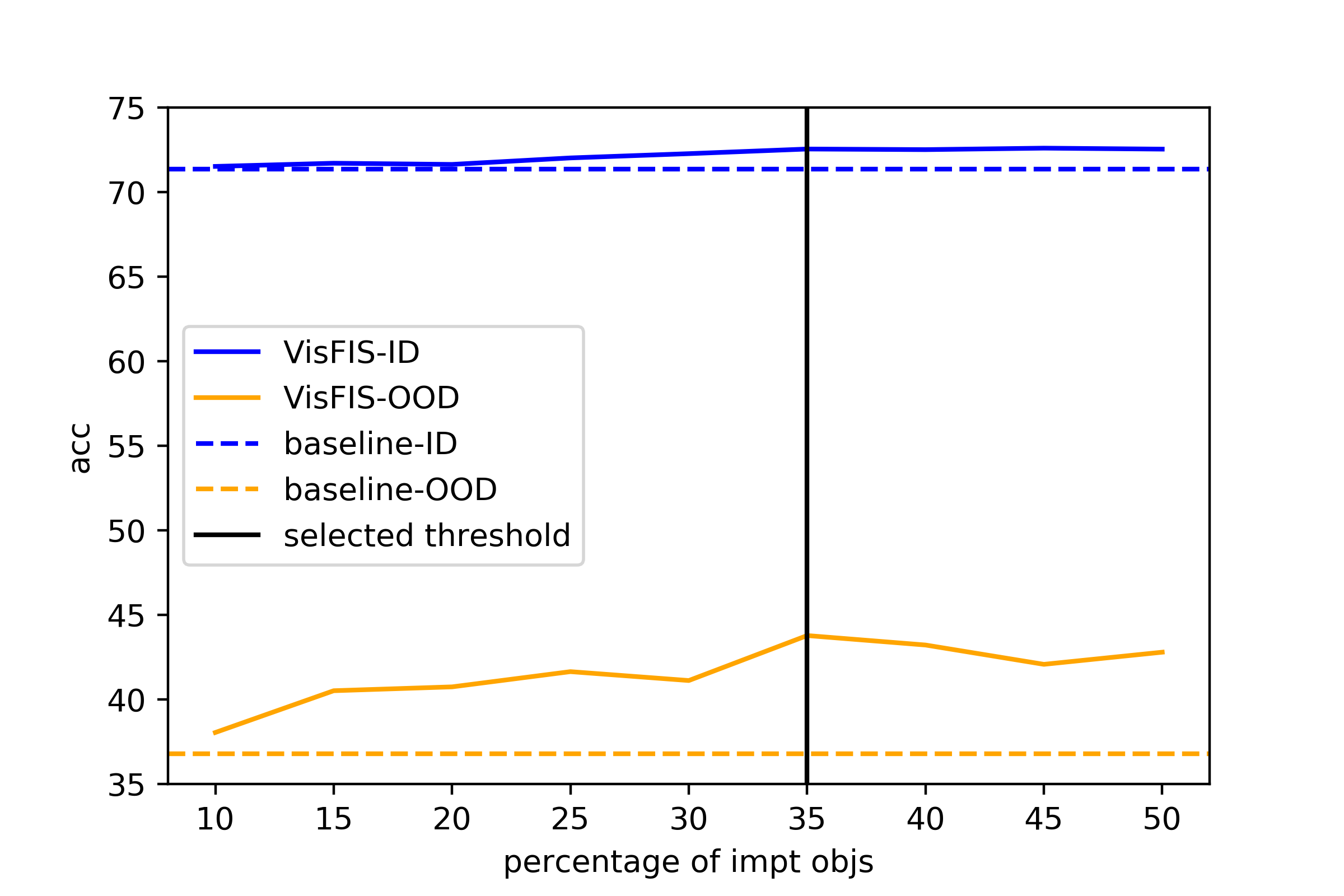}
  \end{center}
  \vspace{-4pt}
  \caption{Threshold ablation on CLEVR-XAI.}
  \label{fig:threshold_tuning}
\end{figure}

\paragraph{Threshold for Human Feature Importance.} For all our objectives except for Align, we need to select the threshold for human FI to separate important features from unimportant ones. We select the threshold separately for each dataset mainly based on (1) qualitative visualizations of the important features and (2) the percentage of data without important features. If a data point is without important features given a threshold, we do not use FI supervision objectives for that datapoint, but we do compute the main task objective, $\mathcal{L}_\textrm{Task}$. 
Although we want the importance features to be reasonable given qualitative visualizations, we don't want to exclude too much data from training. 
We balance between good qualitative results and relatively few excluded data points by selecting thresholds of 0.85, 0.55, and 0.3 for CLEVR-XAI, VQA-HAT, and GQA respectively. These thresholds exclude 1\%, 1\%, and 8\% of data from training with additional objectives for the three datasets. 
To ensure that \method is robust to this choice of threshold, we measure its performance improvement across a range of thresholds using UpDn on CLEVR-XAI. In Fig. \ref{fig:threshold_tuning}, we present model accuracy as a function of the percentage of objects across images that are deemed as important based on a threshold. The values of the threshold vary from 0.1 to 0.98.
The results show that performance improvements in ID and particularly OOD test accuracy are obtainable across a large range of threshold values. 

\begin{figure}
  \vspace{-1pt}
  \begin{center}
    \includegraphics[width=.69\textwidth]{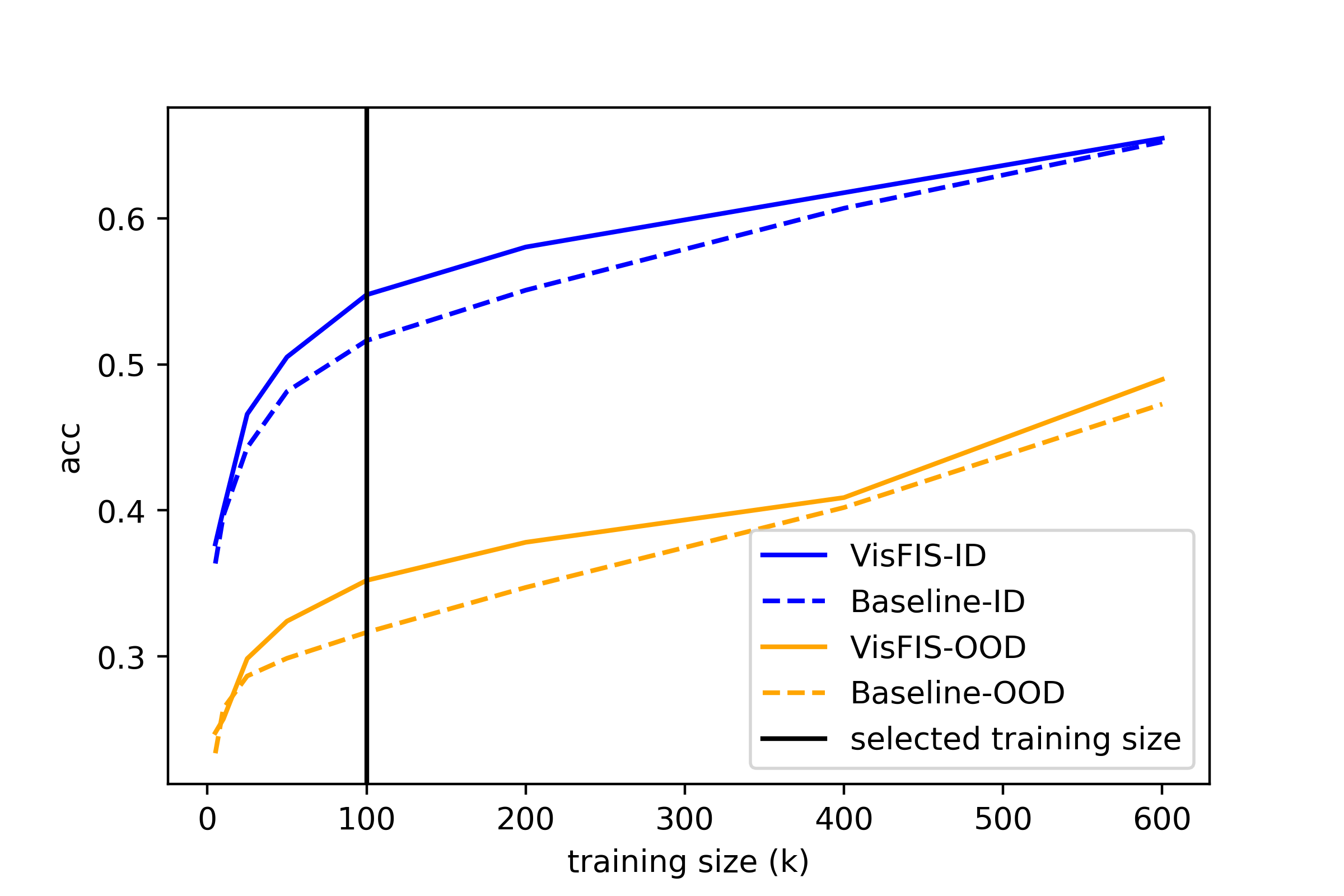}
  \end{center}
  \vspace{-8pt}
  \caption{Training Size Ablation on GQA with an UpDn model.}
  \vspace{-1pt}
  \label{fig:training_size_ablation}
\end{figure}

\paragraph{Training Size Ablation.} GQA dataset \cite{hudson2018gqa} contains 943k training points and 132k validation points. After distribution shift, based on a ratio of 6:1:1.5:1.5, we obtain 645k, 107k, 161k, and 161k data for Train, Dev, Test ID, and Test OOD sets respectively. We then downsample the train set to about 1/6 of its original size. We also exclude a small fraction of data with no ground-truth bounding boxes, and we limit our dev and test sets to 20k points. Thus the final split sizes are 101k train points, 20k Dev, 20k ID Test, and 20k OOD Test. We term this dataset GQA-101k in the main paper.
To measure how FI supervision improvements vary with the amount of training data, we compare \method with the baseline for GQA using between 5k and 600k training points.
Shown in Fig. \ref{fig:training_size_ablation}, the results suggest that supervision is most helpful for improving OOD accuracy when using between 10k and 300k training points, though improvements in OOD accuracy may still be obtained beyond this value.

\paragraph{Categorizing Faithfulness into Worst/Middle/Best Groups.} As part of our analysis of how accuracy varies with explanation plausibility, we group datapoint explanations into three faithfulness categories, Worst, Middle, and Best. We select these based on theoretically sensible values of the Sufficiency and Comprehensiveness metric (see Sec. \ref{sec:metrics} for metric definitions). To be in the Best Sufficiency category, the average Sufficiency score (across explanation sparsity levels) must be at or below 0.01, meaning that the \replaced input must receive a predicted probability no more than one percentage point below the original. For UpDn on CLEVR-XAI, this is about 53\% of the data. To be in the Best Comprehensiveness category, removing the top features must lower the predicted probability by at least 0.4 points (on average across explanation sparsity levels). We give the remaining values and data proportions in Table \ref{tab:explanation_faithfulness_statistics}. 

\begin{table}[t!]
\small
\begin{center}
\caption{Thresholds for categorizing explanation faithfulness and subsequent distribution statistics, for UpDn models on CLEVR-XAI.}
\vspace{-2pt}
\begin{tabular}{l l r r r}
\toprule
& & \multicolumn{3}{c}{Distribution over Faithfulness} \\
\cmidrule(lr){3-5}
Metric & Category & Threshold & Data Proportion \\
    \midrule
    Sufficiency& Worst & $\geq$0.25 & 21\% \\
    Sufficiency& Middle & <0.25 & 25\%  \\
    Sufficiency& Best & <0.01 & 53\% \\
    Comprehensiveness& Worst & <0.20 & 32\% \\
    Comprehensiveness& Middle & <0.40 & 41\%  \\
    Comprehensiveness& Best & $\geq$ 0.40 & 27\% \\
    \bottomrule
\end{tabular}
\label{tab:explanation_faithfulness_statistics}
\end{center}
\vspace{-4pt}
\end{table}

\section{Training Details}
\label{app:training_details}
Our implementations makes use of PyTorch \cite{paszke2017automatic}. Our UpDn model is optimized with a standard Adam \cite{kingma2014adam}, and LXMERT uses Adam with a linear-decayed learning-rate schedule \cite{devlin2019bert}. We use a batch size 64 for UpDn and 32 for LXMERT. For all experiments, we train UpDn for 50 epochs and LXMERT for 35 epochs. UpDn is trained from scratch, while LXMERT uses the default pretrained checkpoint. It takes about an hour to train UpDn on an Nvidia RTX 2080 Ti and about 6 hours for LXMERT on an Nvidia A100.

\paragraph{Hyperparameter Tuning.}
We detail the tuning steps here. All tuning is done using CLEVR-XAI. The tuning is done in sequential order. We first tune learning rate for the baseline UpDn and LXMERT models. Learning rate is chosen from \{1e-2, 5e-3, 1e-3, 5e-4, 1e-4\} for UpDn and \{5e-4, 1e-4, 5e-5, 1e-5\}. We settle with 1e-3 and 5e-5 respectively. We then fix the learning rate and tune the weight $\lambda_i$ for different objectives. For augmentation objectives, we tune the weight with UpDn and use the same weight for LXMERT. The weight for augmentation is chosen from \{100, 10, 1, 1e-1, 1e-2\}, and we end up using weight of 1 for all augmentation objectives. For Inv-FI and Align objectives, we use FI method LOO with all-zeros replacement function, and tune the weight for UpDn and LXMERT separately. For UpDn, the weight is chosen from \{100, 10, 1, 1e-1, 1e-2\}, and for LXMERT, it is chosen from \{1e-3, 1e-4, 1e-5, 1e-6, 1e-7\}. We use weight 1 for UpDn and weight 1e-3 for LXMERT+Inv-FI and weight 1e-6 for LXMERT+Align. We also tune the alignment function - Cosine Similarity, KL divergence, L1 distance, and L2 distance - for the Uncertainty and Align objectives and use KL for Uncertainty and Cosine Similarity for Align. In addition, we tune the weight for HINT \cite{selvaraju2019taking} and SCR \cite{wu2019self} with Vanilla Gradient. The weight is chosen \{10, 1, 1e-1, 1e-2, 1e-3, 1e-4\} for UpDn and \{1e-3, 1e-4, 1e-5, 1e-6, 1e-7\} for LXMERT. We use 1e-3 for UpDn and 1e-6 for LXMERT. We then fix the objective weights and tune the \replace function (results in Table \ref{tab:replace_ablation_table}). Finally, we tune the FI method to use for Inv-FI and Align objectives. KOI-gt works the best with Inv-FI, and Expected Gradient-pred for Align. The numbers for Inv-FI and Align in Table \ref{tab:full_objective_ablation_table} are obtained with KOI-gt and Expected Gradient-pred respectively. We then tune \method with KOI-gt, Expected Gradient-pred, and, for fair comparison with other relevant works, Vanilla Gradient-gt. It turns out that Vanilla Gradient gives the greatest performance gain, and we choose it for all our experiments with \method (see Table \ref{tab:FI_methods_tuning}).

\begin{table}[t!]
\small
\begin{center}
\caption{Feature importance method tuning for \method objective with UpDn model on CLEVR-XAI dev set. The accuracy is averaged over five random seeds. See Appendix \ref{app:training_details} for the full tuning details.}
\vspace{-2pt}
\begin{tabular}{l r}
\toprule
Method & accuracy \\
    \midrule
    Vanilla Grad-gt & 72.55 \\
    KOI-gt & 71.92 \\
    ExpGrad-pred & 72.43 \\
    \bottomrule
\end{tabular}
\label{tab:FI_methods_tuning}
\end{center}
\vspace{-10pt}
\end{table}

\paragraph{Stop Gradient.} When backpropagating through model explanations, we apply a stop gradient for particular FI supervision methods in order to avoid influencing how the model handles the full input (which should be used principally for the task loss $\mathcal{L}_\textrm{Task}$). For FI methods that involve baseline output $f_{\theta}(x)$ or "null" output $f_{\theta}(x_{\vec{0}})$, which includes Excepted Gradient, LOO, KOI, and SHAP, we stop the gradient at $f_{\theta}(x)$ and $f_{\theta}(x_{\vec{0}})$.

\begin{table}[t!]
\small
\begin{center}
\caption{Feature importance method ablation using the Align objective term, for Updn on the CLEVR-XAI dataset. Budget is the number of additional forward and backward passes used by the method.}
\vspace{-2pt}
\begin{tabular}{l r r r r r}
\toprule
& \multicolumn{5}{c}{Accuracy @ Compute Budget} \\
\cmidrule(lr){2-6}
Method & 0 & 1 & 2 & 15 & 30 \\
    \midrule
    Attention & 71.07 & - & - & - & -\\
    Vanilla Grad & - & 71.03 & - & - & - \\
    Expected Grad & - & - & 71.80 & 71.75 & 71.54 \\
    LOO & - & 70.89 & 71.11 & 70.99 & - \\
    KOI & - & - & 71.04 & 71.16 & - \\
    SHAP & - & - & 71.05 & 71.18 & 71.18 \\
    AvgEffect & - & - & 71.05 & 71.03 & 71.15 \\
    \bottomrule
\end{tabular}
\label{tab:FI_method_ablation_table}
\end{center}
\vspace{-10pt}
\end{table}

\begin{table}[t!]
\small
\begin{center}
\caption{Objective term ablation for the CLEVR-XAI dataset with an UpDn model}
\vspace{-2pt}
\begin{tabular}{l r r r r r r r r}
\toprule
& \multicolumn{2}{c}{Accuracy} & \multicolumn{3}{c}{RRR Metrics} & \multicolumn{3}{c}{Expl. Metrics} \\
\cmidrule(lr){2-3} \cmidrule(lr){4-6}  \cmidrule(lr){7-9}
Objective & \multicolumn{1}{c}{ID $\uparrow$} & \multicolumn{1}{c}{OOD $\uparrow$} & \multicolumn{1}{c}{Suff $\uparrow$} & \multicolumn{1}{c}{Inv $\uparrow$} & \multicolumn{1}{c}{Unc $\downarrow$} & \multicolumn{1}{c}{Plau} & \multicolumn{1}{c}{Suff $\downarrow$} & \multicolumn{1}{c}{Comp $\uparrow$} \\
    \midrule
    Baseline & 71.37 & 36.80 & 48.82 & 77.89 & 55.17 & 28.82 & 34.66 & 47.56 \\
    Saliency Guided & 71.50 & 37.71 & 73.00 & 92.17 & 76.98 & 13.84 & -7.13 & 21.23\\
    Inv-DA & 71.17 & 35.91 & 72.53 & \textbf{93.12} & 76.29 & 14.33 & \textbf{-7.32} & 21.30 \\
    Inv-FI & 71.41 & 38.88 & 45.31 & 76.34 & 71.41 & 28.60 & 35.79 & \textbf{48.20} \\
    Uncertainty & 71.30 & 38.34 & 10.75 & 86.58 & \textbf{4.16} & 8.56 & 73.49 & 41.65 \\
    Align & 72.04 & 41.61 & 61.19 & 79.51 & 64.22 & 37.20 & 26.86 & 35.18 \\
    Suff-Random & 71.73 & 39.08 & 73.59 & 92.59 & 60.93 & 17.32 & -5.29 & 22.48 \\
    Suff-Human & 71.87 & 40.91 & 76.94 & 90.82 & 81.42 & 16.27 & -6.68 & 26.45 \\
    + Align & 72.42 & 41.63 & \textbf{78.55} & 89.69 & 80.02 & \textbf{35.73} & 0.53 & 27.18 \\
    + Unc & 72.33 & 41.54 & 77.83 & 89.70 & 41.68 & 23.41 & -5.18 & 37.15 \\
    + Align+Unc+Inv & \textbf{72.82} & \textbf{43.78} & 76.65 & 91.72 & 43.64 & 22.67 & -0.30 & 29.51 \\
    \bottomrule
\end{tabular}
\label{tab:full_objective_ablation_table}
\end{center}
\vspace{-4pt}
\end{table}

\section{Additional Results}
\label{app:additional_results}

\begin{figure}[t]
  \vspace{-4pt}
  \begin{center}
    \includegraphics[width=.81\textwidth]{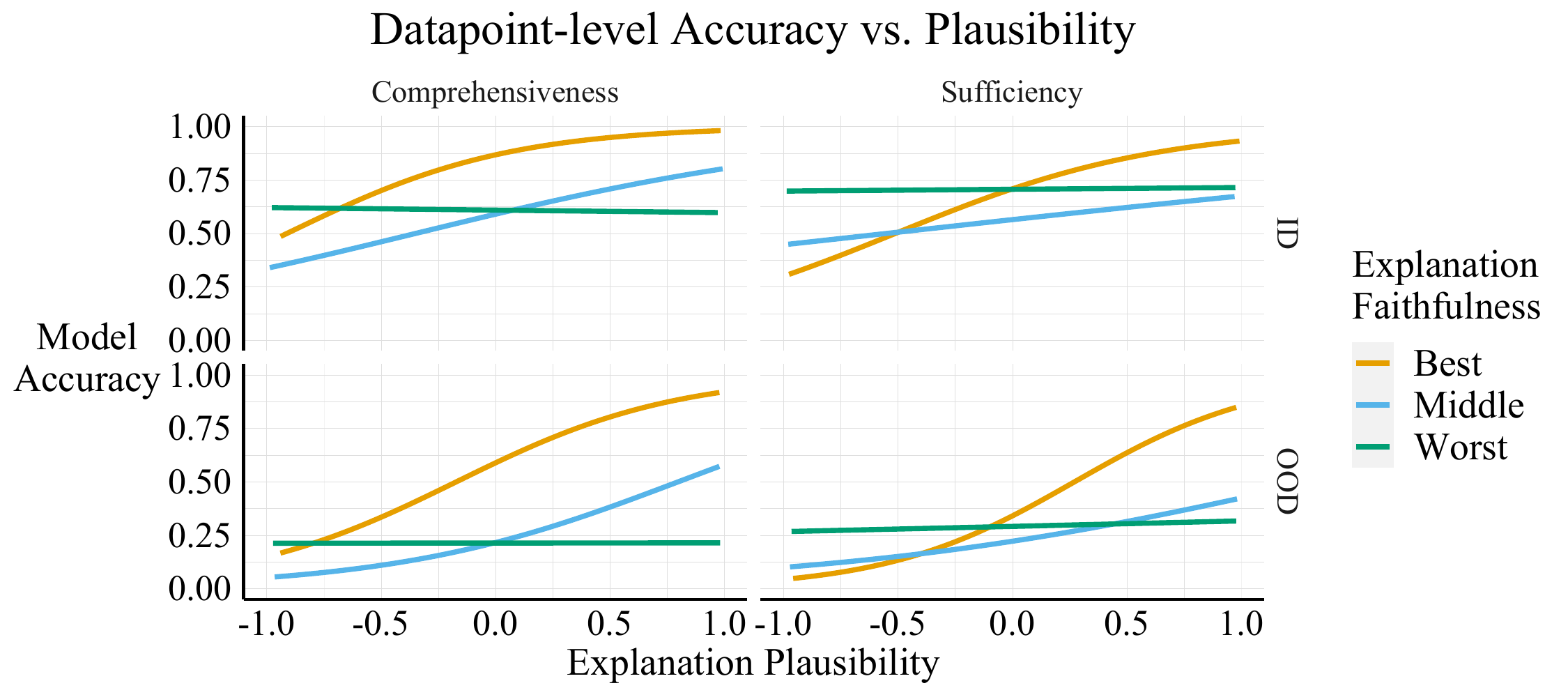}
  \end{center}
  \vspace{-5pt}
  \caption{
  Datapoint level accuracy by explanation plausibility and faithfulness, for CLEVR-XAI models, grouped by faithfulness metric and test split.
  }
  \label{fig:acc_by_plau_x_faith_all}
  \vspace{-2pt}
\end{figure}

\begin{figure}[t]
  \vspace{-4pt}
  \begin{center}
    \includegraphics[width=.93\textwidth]{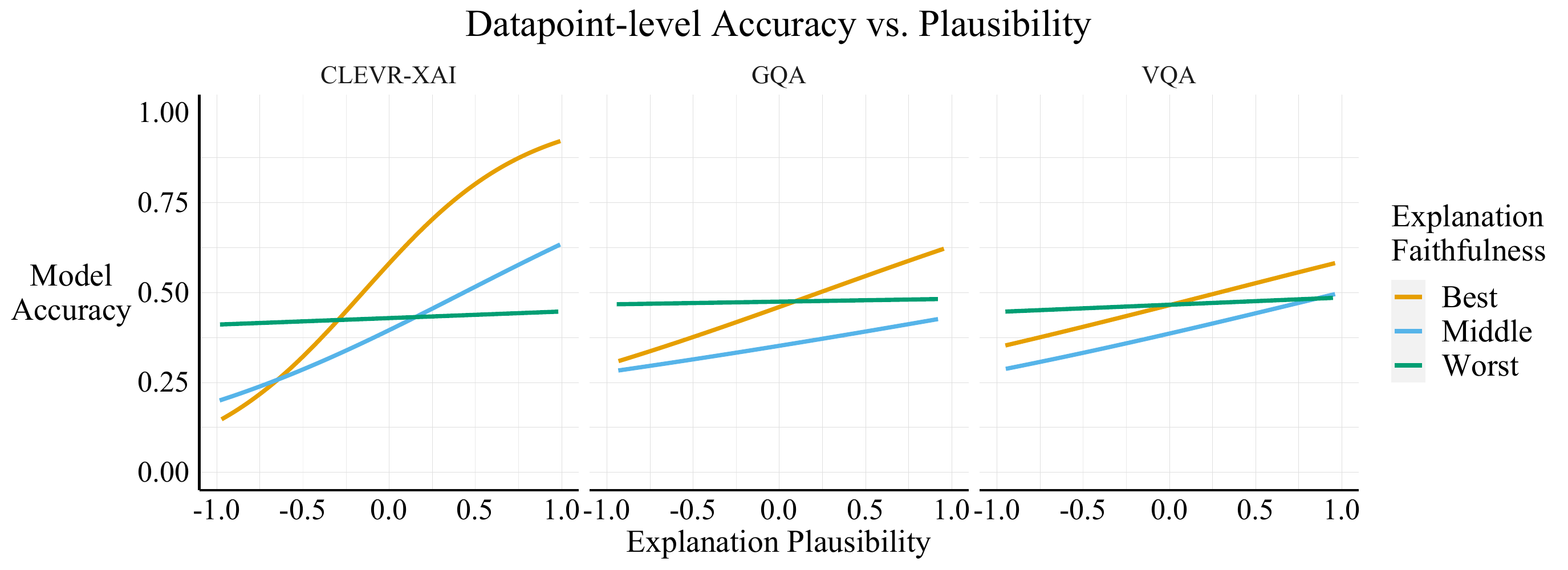}
  \end{center}
  \vspace{-5pt}
  \caption{
  Datapoint level accuracy by explanation plausibility and faithfulness for UpDn models, grouped by dataset.
  }
  \label{fig:acc_by_plau_x_faith_dataset}
  \vspace{-2pt}
\end{figure}

\begin{figure}[t]
  \vspace{-4pt}
  \begin{center}
    \includegraphics[width=.51\textwidth]{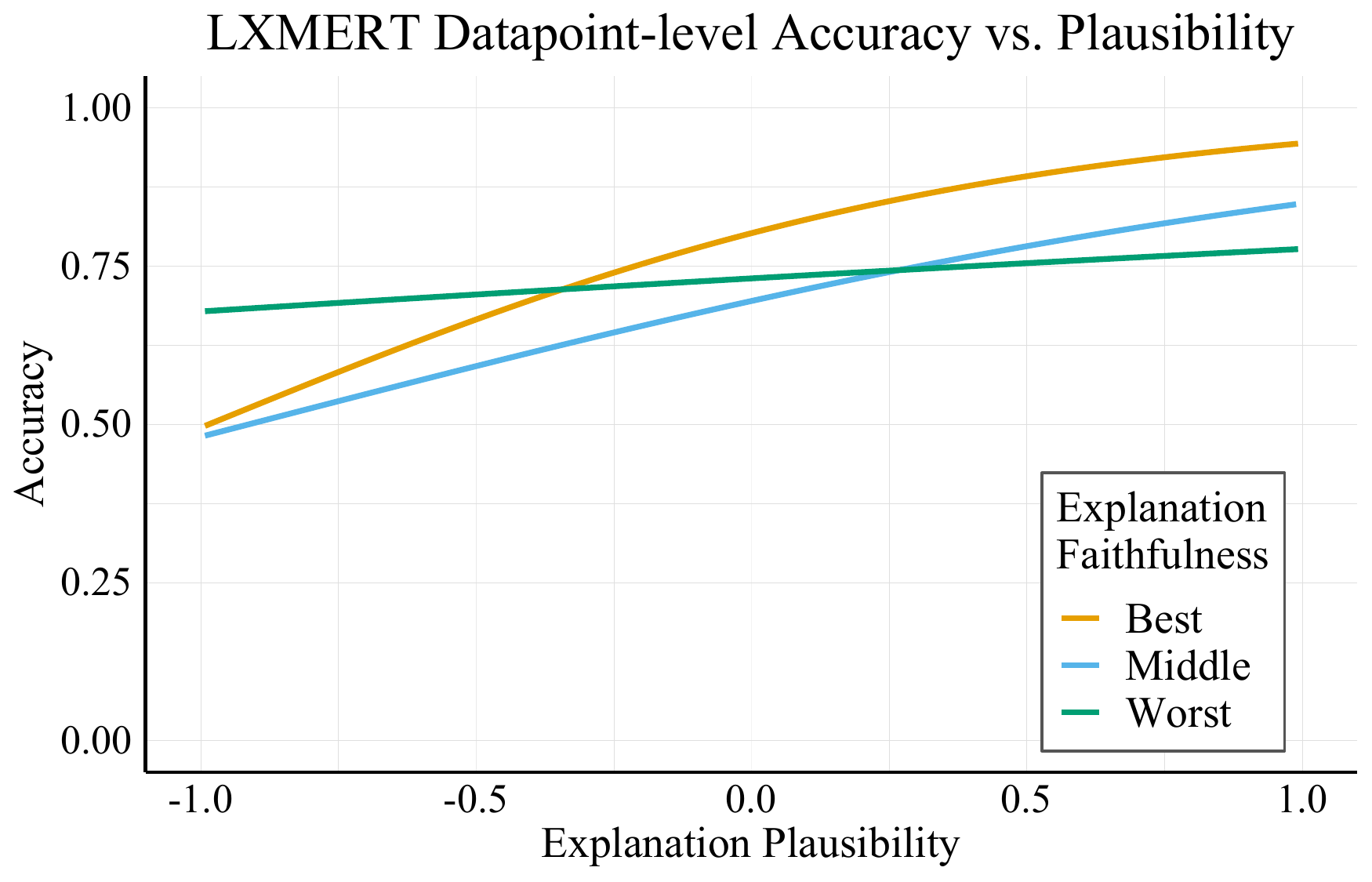}
  \end{center}
  \vspace{-5pt}
  \caption{
  Datapoint level accuracy by explanation plausibility and faithfulness, for LXMERT on CLEVR-XAI, averaged across faithfulness metrics and test splits.
  }
  \label{fig:acc_by_plau_x_faith_single_lxmert}
  \vspace{-2pt}
\end{figure}

\subsection{Which Objectives Are Affected by Random Supervision?}

\textbf{Design.} In earlier experiments, we find that \method does not improve performance with random supervision. Here, we further explore how each of the four additional objective terms in \method is individually influenced by random supervision.
To assess the effect of random supervision on each objective, we give random supervision to one of the objectives and normal supervision to the other three on CLEVR-XAI with UpDn.

\textbf{Results.} We show the results in Table \ref{tab:random_sup}. The Suff-Human objective is the main reason why \method does not work with random supervision. Uncertainty and alignment objectives with random supervision hurt the performance, but not as much as the sufficiency objective. Note that Suff-Human with random supervision is different from Suff-Random, which has different features mask out across the training process for the same sample. Here, Suff-Human with random supervision has the same (random) features masked out for the entire training process. The invariance objective with random supervision does not hurt the performance at all. 

\begin{figure}
  \vspace{-4pt}
  \begin{center}
    \includegraphics[width=.97\textwidth]{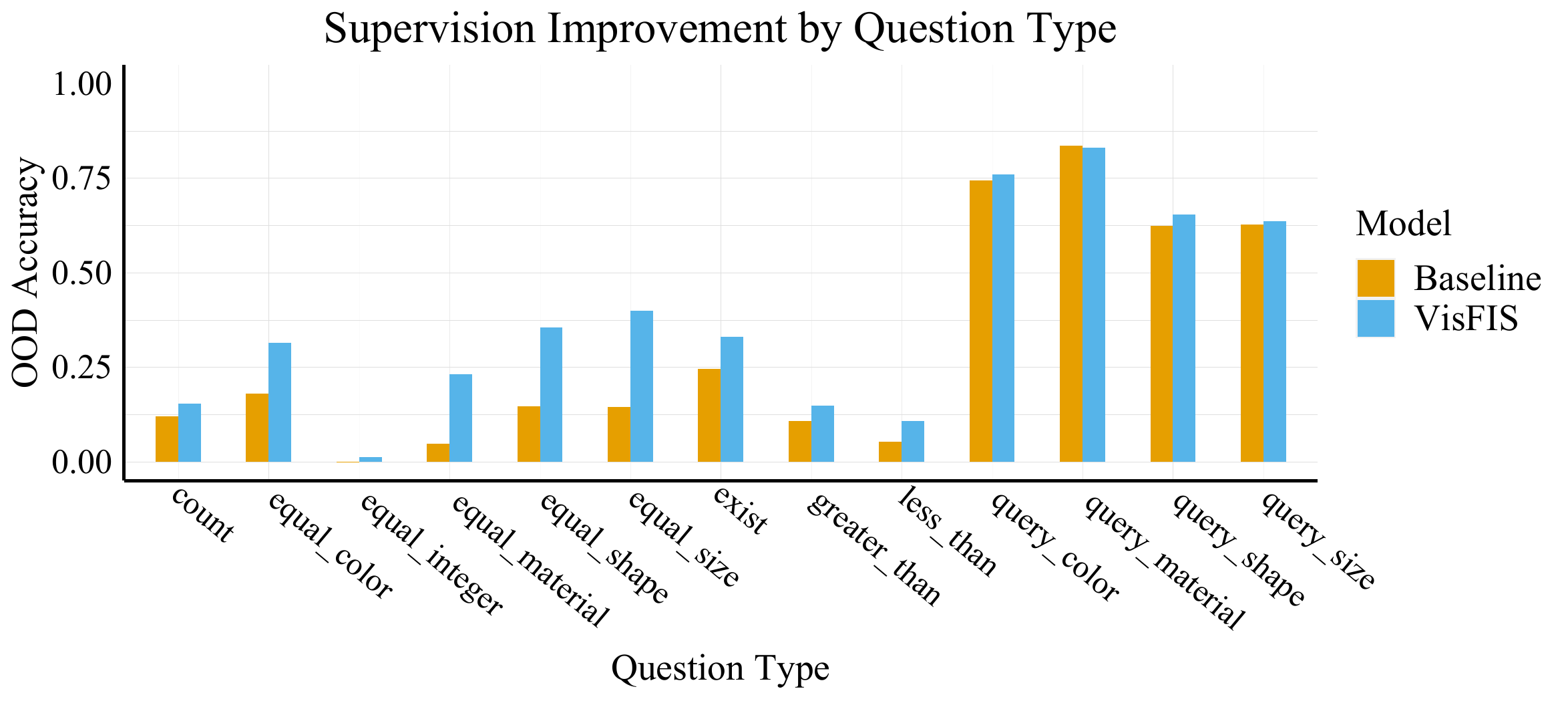}
  \end{center}
  \vspace{-5pt}
  \caption{OOD accuracy for the baseline and \method on CLEVR-XAI with UpDn, grouped by question type.}
  \label{fig:xai_barplot}
  \vspace{-3pt}
\end{figure}

\begin{table}[t!]
\small
\begin{center}
\caption{Random supervision control experiments on UpDn + CLEVR-XAI for different objective terms in \method. We use a fixed set of random explanations for one objective at a time.}
\vspace{-2pt}
\begin{tabular}{l l l}
\toprule
Method & ID acc & OOD acc \\
    \midrule
    Baseline & 71.30 & 36.80 \\
    \method & 72.82 & 43.78 \\
    w/ random Suff-Human & 69.93 & 36.70 \\
    w/ random Unc & 72.51 & 41.87 \\
    w/ random Align & 71.27 & 39.58 \\
    w/ random Inv-FI & 72.59 & 44.20 \\
    \bottomrule
\end{tabular}
\label{tab:random_sup}
\end{center}
\vspace{-10pt}
\end{table}

\subsection{Accuracy-Plausibility Relationship Across Test Splits, Datasets, and Models}

In the main paper Fig. \ref{fig:acc_by_plau}, we show how accuracy varies as a function of explanation plausibility and faithfulness for UpDn models on CLEVR-XAI, and we group data points across ID and OOD test splits. Here, we show that the main trends are generally consistent across the choice of explanation metric (Sufficency vs. Comprehensiveness), test split (ID vs. OOD), dataset, and model.
Trends across metric and split are shown in Fig. \ref{fig:acc_by_plau_x_faith_all}, and trends across datasets are shown in Fig. \ref{fig:acc_by_plau_x_faith_dataset}. We show results for LXMERT on CLEVR-XAI in Fig. \ref{fig:acc_by_plau_x_faith_single_lxmert}.
Though the trends weaken slightly in certain settings, we always find that accuracy correlates positively with plausibility for highly faithful explanations, while the relationship is weaker or non-existent for unfaithful explanations.

\subsection{Which FI Method Produces the Most Faithful Explanations?}

\textbf{Design.} We calculate the Explanation Sufficiency and Explanation Comprehensiveness metrics for the FI methods listed in Appendix \ref{app:explanation_methods}, using either the predicted or ground truth class to select the output logit that is explained. All experiments are conducted on CLEVR-XAI with UpDn. Following guidelines from \citet{hase2021out}, the UpDn models are trained with Suff-Random objective to make the replaced features in-distribution for the models. 
For LOO and KOI, we use a budget of 15 and 36 on CLEVR-XAI and VQA-HAT/GQA, which is the same number as the number of bounding boxes. For SHAP, Average Effect, and Expected Gradient, we use a budget of 1000 to reduce noise in deriving each explanation, as these methods involve random sampling. We select the best explanation method for each dataset by taking the best score on average across the two metrics.

\textbf{Results.} We show the results in Table \ref{tab:FI_metrics}. In general, explanations obtained on predicted class are more faithful to the model decisions than those obtained on ground truth class. UpDn attention, LOO-pred, and KOI-pred are among the best across three datasets. SHAP and Average Effect surprisingly are not very faithful across all three datasets. KOI on predicted class is the most faithful one for CLEVR-XAI and VQA-HAT, while LOO on predicted class is the best for GQA. Hence, when calculating explanation metrics, we use KOI on predicted class for CLEVR-XAI and VQA-HAT and LOO on predicted class for GQA.

\begingroup
\setlength{\tabcolsep}{5pt}
\begin{table}[t!]
\small
\begin{center}
\caption{FI tuning for explanation metrics with UpDn models on Dev ID data.}
\vspace{-2pt}
\begin{tabular}{l r r r r r r}
\toprule
& \multicolumn{2}{c}{CLEVR-XAI} & \multicolumn{2}{c}{VQA-HAT} & \multicolumn{2}{c}{GQA-101k} \\
\cmidrule(lr){2-3} \cmidrule(lr){4-5}  \cmidrule(lr){6-7}
FI Method & \multicolumn{1}{c}{Suff $\downarrow$} & \multicolumn{1}{c}{Comp $\uparrow$} & \multicolumn{1}{c}{Suff $\downarrow$} & \multicolumn{1}{c}{Comp $\uparrow$} & \multicolumn{1}{c}{Suff $\downarrow$} & \multicolumn{1}{c}{Comp $\uparrow$} \\
    \midrule
    UpDn Attention & 1.19±0.30 & 20.46±0.72 & 4.08±4.48 & 9.06±2.83 & 0.08±0.40 & \textbf{11.29±1.65} \\
    Vanilla Grad-pred & 8.80±1.79 & 12.70±1.20 & 8.06±4.24 & 5.70±4.33 & 14.76±1.67 & 1.60±1.23 \\
    Vanilla Grad-gt & 5.04±1.08 & 16.01±0.83 & 13.21±2.32 & 5.46±3.44 & 15.64±2.27 & 3.00±0.97 \\
    ExpGrad-pred & 5.15±1.92 & 12.39±2.05 & 3.41±5.06 & 9.20±2.91 & 0.25±0.38 & 7.90±1.80 \\
    ExpGrad-gt & 9.71±1.12 & 9.78±1.58 & 6.12±3.85 & 6.91±3.46 & 5.00±1.22 &4.43±0.99 \\
    LOO-pred & -5.01±0.24 & 14.34±0.82 & -2.51±6.97 & 9.86±3.99 & \textbf{-3.32±0.33} & 9.26±1.82 \\
    LOO-gt & 2.62±0.73 & 9.67±0.48 & 5.75±4.04 & 4.35±1.82 & 5.31±1.62 & 3.14±0.82 \\
    KOI-pred & \textbf{-5.17±0.32} & \textbf{22.43±1.14} & \textbf{-3.45±7.34} & \textbf{10.61±4.31} & 6.05±7.61 & 7.81±2.68 \\
    KOI-gt & 3.25±0.73 & 18.23±0.77 & 7.71±3.57 & 5.40±2.28 & 19.63±6.27 & 3.62±1.24 \\
    SHAP-pred & 17.06±0.73 & 2.86±0.19 & 36.12±12.57 & 2.20±3.03 & 15.25±7.98 & 0.04±0.14 \\
    SHAP-gt & 17.07±0.76 & 2.84±0.19 & 33.87±12.60 & 2.11±2.82 & 13.53±7.07 & 0.04±0.16 \\
    Average Effect-pred & 17.35±0.71 & 2.83±0.18 & 7.51±3.30 & 3.79±4.58 & 1.69±0.14 & 0.14±0.22 \\
    Average Effect-gt & 17.46±0.72 & 2.74±0.20 & 7.38±3.37 & 3.88±4.52 & 1.68±0.14 & 0.14±0.22 \\
    \bottomrule
\end{tabular}
\label{tab:FI_metrics}
\end{center}
\vspace{-1pt}
\end{table}
\endgroup

\subsection{Can Explanation Supervision Improve Model Explainability?}

\looseness=-1
\textbf{Design.} To assess the effect of FI supervision on model explainability, we record faithfulness metrics using all of our CLEVR-XAI models. We then plot explanation Sufficiency and Comprehensiveness for each model (averaged across five seeds) to visualize the distribution of faithfulness scores. 

\begin{figure}
  \vspace{-4pt}
  \begin{center}
    \includegraphics[width=.79\textwidth]{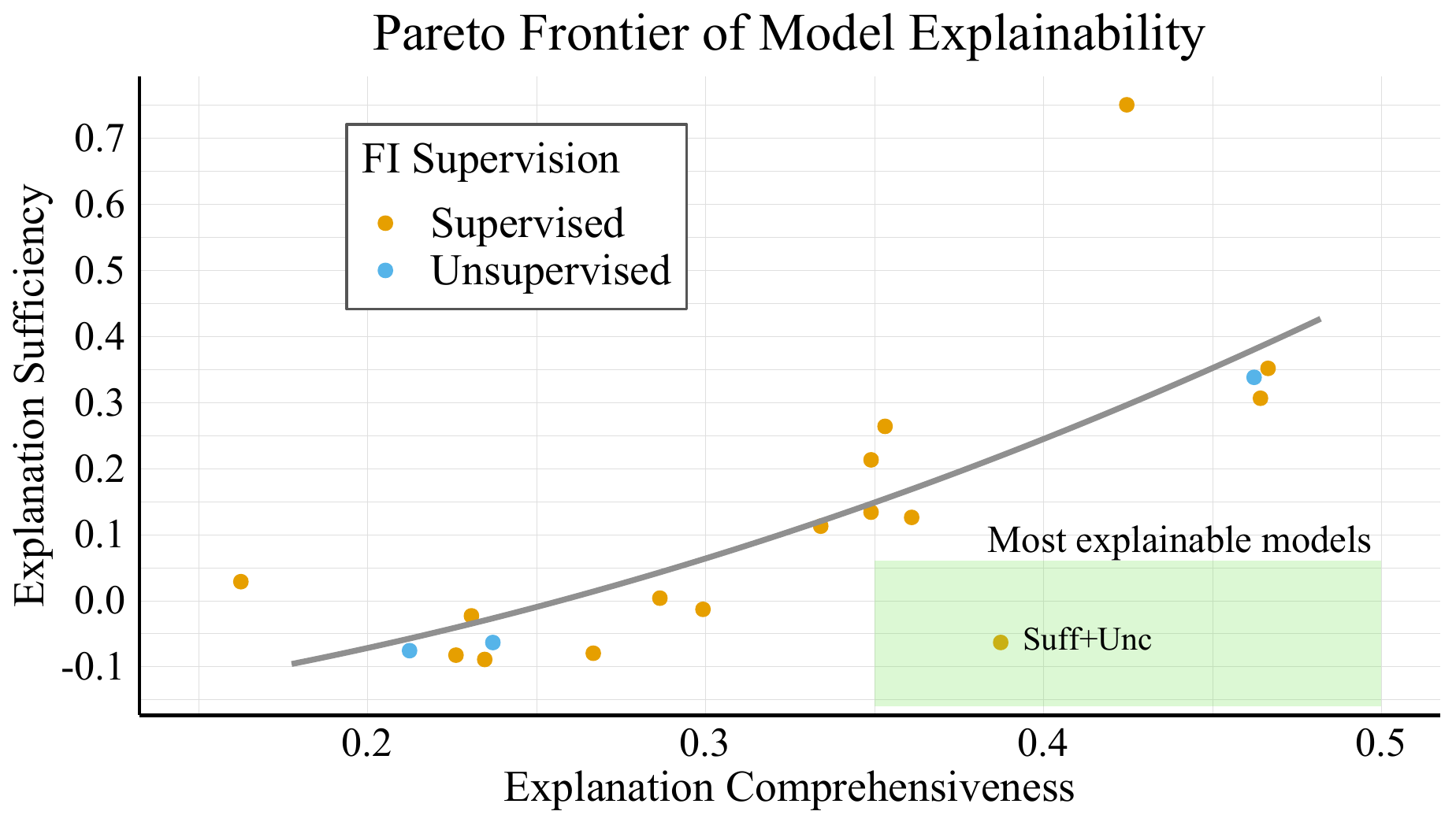}
  \end{center}
  \vspace{-5pt}
  \caption{Average model explanation Sufficiency and Comprehensiveness scores (shown for models on in-distribution CLEVR data). }
  \label{fig:pareto}
  \vspace{-5pt}
\end{figure}

\textbf{Results.} We show results for each model in Fig. \ref{fig:pareto} (scores also listed in Appendix Table \ref{tab:full_objective_ablation_table}). 
We find that average explanation Sufficiency and Comprehensiveness scores lie along a Pareto frontier, shown by the gray line, which represents a trade-off between better Sufficiency and Comprehensiveness (models better in one metric are worse in the other). 
Generally, explanation supervision does not improve model explainability relative to unsupervised models, with the exception of the Suff+Unc objective. In the bottom right of the plot, this model demonstrates a better combination of Sufficiency and Comprehensiveness than other supervised or unsupervised methods, including Saliency-Guided Training \cite{ismail2021improving}. The Suff+Unc model is especially explainable likely because the Sufficiency objective encourages the model to rely on a small number of important features, while the Uncertainty objective encourages the model to become less confident when those important features are removed.

\begin{table}[t!]
\small
\begin{center}
\caption{Datapoint level faithfulness distributions (in terms of Sufficiency) conditional on datapoint-level and model-level plausibility scores, averaged across CLEVR-XAI models. }
\vspace{-2pt}
\begin{tabular}{l l r r r}
\toprule
& & \multicolumn{3}{c}{Distribution over Faithfulness} \\
\cmidrule(lr){3-5}
Model Plausibility & Data Plausibility & Worst & Medium & Best \\
    \midrule
Low & Low & 0.51 & 0.27 & 0.22 \\
Low & Middle & 0.19 & 0.41 & 0.40 \\
Low & High & 0.11 & 0.49 & 0.40 \\
Middle & Low & 0.02 & 0.13 & 0.85 \\
Middle & Middle & 0.01 & 0.1 & 0.89 \\
Middle & High & 0.01 & 0.07 & 0.92 \\
High & Low & 0.24 & 0.31 & 0.45 \\
High & Middle & 0.20 & 0.27 & 0.52 \\
High & High & 0.18 & 0.23 & 0.60\\
    \bottomrule
\end{tabular}
\label{tab:conditional_distribution_table}
\end{center}
\vspace{-4pt}
\end{table}

\subsection{How Can Models with Low Plausibility Achieve High Accuracy?} 
Shown in Table \ref{tab:conditional_distribution_table}, we find that models with lower average plausibility show different conditional relationships than models with higher average plausibility, which helps explain why low-average-plausibility models can achieve similar accuracies to high-average-plausibility models. Low-average-plausibility models have low plausibility points with low faithfulness scores, meaning these points are still often accurately predicted and hence do not bring down the average model accuracy. Meanwhile, middle and high-average-plausibility models often have low-plausibility points with highly faithful explanations, meaning these points are often inaccurately predicted, offsetting any gains to average model accuracy that are achieved for points with both highly plausible and faithful explanations.

\begingroup
\begin{table}[ht!]
\small
\begin{center}
\vspace{-1pt}
\caption{Test accuracy for Updn model on full VQA test set, including all question types.}
\vspace{-5pt}
\begin{tabular}{l r r r r r r}
\toprule
& \multicolumn{2}{c}{VQA-HAT} \\
\cmidrule(lr){2-3}
Method & \multicolumn{1}{c}{ID} & \multicolumn{1}{c}{OOD} \\
\midrule
    Baseline & 52.22 ± 0.92 & 38.95 ± 0.91\\
    Suff-Random & 52.26 ± 0.90 & 39.30 ± 0.97 \\
    \citet{selvaraju2019taking} & 52.11 ± 1.01 & 37.95 ± 1.07 \\
    \citet{wu2019self} & 52.16 ± 0.94 & 38.53 ± 0.94 \\
    \citet{simpson2019gradmask} & 52.32 ± 0.91 & 38.84 ± 1.08 \\
    \citet{chang2021towards} & 50.42 ± 1.01 & 31.29 ± 1.44\\
    \citet{singla2022core} &  52.93 ± 0.96 & 39.05 ± 1.64\\
    \method & 52.79 ± 0.95 & 40.49 ± 0.96 \\
    \midrule 
    \ w/ Rand. Supervis. & 52.21 ± 0.94 & 37.95 ± 0.99 \\
    \bottomrule
\end{tabular}
\label{tab:full_vqa_table}
\end{center}
\vspace{-9pt}
\end{table}
\endgroup

\subsection{Do RRR Metrics Predict OOD Generalization? Additional Datasets and Models}

We measure the correlation between RRR metrics (calculated with ID data) and OOD accuracy across a large set of models. We report results additional in Table \ref{tab:metric_table_others} here for LXMERT models on CLEVR-XAI and UpDn for GQA/VQA.
We perform a cross-validation resampling model-level statistics 10k times, using 40 models' metrics as training data and 5 for testing each time. The final metrics we consider are: (1) ID accuracy on its own as a baseline, (2) RRR metrics on their own, (3) ID accuracy plus average model confidence, (4) ID accuracy plus explanation metrics, (5) ID accuracy plus RRR metrics, and (6) All Metrics, which uses all available metrics. The results are similar to in the main paper, showing that RRR metrics do not achieve a better correlation with OOD accuracy than ID accuracy does on its own.

\begin{table}
\small
\vspace{-1pt}
\begin{center}
\caption{Correlations between metrics and OOD accuracy for additional datasets and model architectures. We derive results from 45 models (differing by seed and objective) per condition.}
\vspace{2pt}
\begin{tabular}{l r r r r r r}
\toprule
& \multicolumn{2}{c}{UpDn + VQA-HAT} & \multicolumn{2}{c}{UpDn + GQA-101k} & \multicolumn{2}{c}{LXMERT + CLEVR-XAI}\\
\cmidrule(lr){2-3} \cmidrule(lr){4-5} \cmidrule(lr){6-7}
Metric & \multicolumn{1}{c}{Train} & \multicolumn{1}{r}{Test} & \multicolumn{1}{r}{Train} & \multicolumn{1}{r}{Test} & \multicolumn{1}{r}{Train} & \multicolumn{1}{r}{Test} \\
    \midrule
    RRR-Suff & 0.393 & 0.627 & 0.644 & 0.584 & 0.464 & 0.553 \\
    RRR-Inv & 0.011 & 0.148 & 0.549 & 0.526 & 0.035 & 0.160 \\
    RRR-Unc & 0.470 & 0.530 & 0.478 & 0.459 & -0.111 & 0.024 \\
    ID Acc & 0.952 & 0.850 & 0.908 & 0.859 & 0.903 & \textbf{0.898} \\
    + Model Conf. & 0.957 & \textbf{0.866} & 0.921 & \textbf{0.876} & 0.910 & 0.858 \\
    + Expl. Metrics & 0.956 & 0.847 & 0.923 & 0.873 & 0.923 & 0.883 \\
    + RRR-all & 0.958 & 0.846 & 0.929 & 0.875 & 0.920 & 0.859 \\
    All Metrics & \textbf{0.965} & 0.816 & \textbf{0.943} & 0.832 & \textbf{0.938} & 0.768  \\
    \bottomrule
\end{tabular}
\label{tab:metric_table_others}
\end{center}
\vspace{-9pt}
\end{table}

\end{document}